\documentclass[sigconf]{acmart}

\copyrightyear{2022}
\acmYear{2022}
\setcopyright{othergov}\acmConference[ICPP '22]{51st International Conference on Parallel Processing}{August 29-September 1, 2022}{Bordeaux, France}
\acmBooktitle{51st International Conference on Parallel Processing (ICPP '22), August 29-September 1, 2022, Bordeaux, France}
\acmPrice{15.00}
\acmDOI{10.1145/3545008.3545085}
\acmISBN{978-1-4503-9733-9/22/08}

\usepackage{url}
\usepackage{epstopdf}
\usepackage{xspace}
\usepackage{amsmath}
\usepackage{subfigure}
\usepackage{multirow}
\usepackage{xcolor}

\usepackage[ruled,linesnumbered]{algorithm2e}
\SetKw{KwDownTo}{downto}
\SetKw{KwTo}{to}
\SetKw{KwTrue}{true}
\SetKw{KwFalse}{false}
\SetKwInOut{Input}{Input}
\SetKwInOut{Output}{Output}
\SetKw{KwAnd}{and}
\SetKw{KwBreak}{break}

\usepackage{algorithmic}
\usepackage{amsthm}
\renewcommand{\algorithmicrequire}{\textbf{Input:}}
\renewcommand{\algorithmicensure}{\textbf{Output:}}

\renewcommand{\algorithmiccomment}[1]{// #1}

\usepackage{enumitem}
\usepackage{colortbl}

\newcommand{\revision}[1]{{\color{black}{#1}}}
\begin{document}

\title[]{FedDRL: Deep Reinforcement Learning-based Adaptive Aggregation for Non-IID Data in Federated Learning}

\author{Nang Hung Nguyen}
\authornote{Both authors contributed equally to the paper}
\author{Phi Le Nguyen}
\authornotemark[1]
\email{hung.nn184118@sis.hust.edu.vn}
\email{lenp@soict.hust.edu.vn}
\affiliation{
  \institution{Hanoi University of Science and}            
  \city{Technology, Hanoi}
  \country{Vietnam}                    
}

\author{Duc Long Nguyen}
\author{Trung Thanh Nguyen}
\email{long.nd183583@sis.hust.edu.vn} 
\email{thanh.nt176874@sis.hust.edu.vn} 
\affiliation{
  \institution{Hanoi University of Science and}            
  \city{Technology, Hanoi}
  \country{Vietnam}                    
}

\author{Thuy Dung Nguyen}
\author{Thanh Hung Nguyen}
\email{dung.nt184244@sis.hust.edu.vn} 
\email{hungnt@soict.hust.edu.vn}
\affiliation{
  \institution{Hanoi University of Science and}            
  \city{Technology, Hanoi}
  \country{Vietnam}                    
}

\author{Huy Hieu Pham}
\email{hieu.ph@vinuni.edu.vn} 
\affiliation{
  \institution{College of Engineering \& Computer Science and VinUni-Illinois Smart Health Center, VinUniversity} 
  \email{hieu.ph@vinuni.edu.vn}
  \city{Hanoi}
  \country{Vietnam}                    
}

\author{Truong Thao Nguyen}
\affiliation{
  \institution{National Institute of Advanced Industrial Science and Technology}            
  \city{Tokyo}
  \country{Japan}                    
}
\email{nguyen.truong@aist.go.jp}          

\renewcommand{\shortauthors}{Nang Hung Nguyen et al.}

\begin{abstract}
The uneven distribution of local data across different edge devices (clients) results in slow model training and accuracy reduction in federated learning. Naive federated learning (FL) strategy and most alternative solutions attempted to achieve more fairness by weighted aggregating deep learning models across clients.
This work introduces a novel non-IID type encountered in real-world datasets, namely cluster-skew, in which groups of clients have local data with similar distributions, causing the global model to converge to an over-fitted solution.  
To deal with non-IID data, particularly the cluster-skewed data, we propose FedDRL, a novel FL model that employs deep reinforcement learning to adaptively determine each client's impact factor (which will be used as the weights in the aggregation process).
Extensive experiments on a suite of federated datasets confirm that the proposed FedDRL improves favorably against FedAvg and FedProx methods, e.g.,  up to $4.05$\% and $2.17$\% on average for the CIFAR-100 dataset, respectively.
\end{abstract}


\maketitle
\section{Introduction}
\label{sec:intro}

In the last few years, the number of mobile and IoT devices, as well as their computing powers, has been rapidly growing. Along with the development of advanced machine learning techniques and the increasing privacy concerns, Federated Learning (FL)~\cite{fedavg_mcmahan2017communication,zhu2021federated} has become one of the main paradigms where deep learning (DL) models are trained in parallel at a large number of edge devices (clients).
Unlike the de facto conventional machine learning at a centralized supercomputer and data center~\cite{ben2018demystifying}, FL does not require to collect the private raw data of clients. 
Instead, FL retains such data locally and never uploads to a server, yet allows the clients to gather the benefit of rich data across all the clients by sharing a global DL model aggregated at the server.
Specifically, the clients perform the local training using the latest global model in several local steps (epochs) on their local data set. Each client then sends its locally trained model, usually the gradients or weights of the DL model, to the server.
In a communication round, after collecting all the locally trained models from all the clients, the server calculates the new global model, e.g., by taking the average of all the locally trained models~\cite{fedavg_mcmahan2017communication}.

In principle,  one of the advantages of FL is the guarantee in training a DL model over privacy-sensitive data for a wide range of real-world applications. 
However, one of the key challenges which limit the usage of FL settings is the statistical heterogeneity~\cite{fedprox_li2020federated}.
FL involves a large number of edge devices/clients. The distribution of local data across clients is significantly different since the training data generated at a given client typically depends on how this particular edge mobile device is used~\cite{fedavg_mcmahan2017communication}. Therefore, the local dataset of a client may not reflect the global distribution, yet it is non-independent and identical data across the edge devices (\textbf{non-IID} data)~\cite{zhu2021federated,li2019convergence,IWQOS2021_fedacs,icml2020_hsieh20a_skewscout}. 
It has been shown that standard FL methods such as FedAvg~\cite{fedavg_mcmahan2017communication} are not well-designed to address the challenges of non-IID data. It could increase the training time and degrade the training accuracy in practice~\cite{li2019convergence,xiao2020averaging,fednova,li2020federated}.

To tackle the statistical heterogeneity problem, many efforts focused on actively selecting the clients for balancing the data distribution~\cite{cho2020client,favor_hwangInforcom2020,inforcom2021_fedsens}. Alternatively, many studies improved the loss function by devoting the optimization techniques ~\cite{fedprox_li2020federated, fednova,ICPP2021_fedCav}.
Another approach tackles the issue of unfairness when performing the weight aggregation at the server, i.e., overly favoring the DL models on some devices either with a higher number of samples~\cite{fedadp} or with significant differences in the magnitudes of conflict gradients~\cite{ijcai2021_Fairness_fedfv}.
It is stated that in the standard FL methods such as FedAvg~\cite{fedavg_mcmahan2017communication}, the average accuracy may be high, yet the accuracy of each device may not always be very good~\cite{fedfa_huang2020fairness}. Therefore, several works design weighting strategies to encourage a more fair aggregation method~\cite{ijcai2021_Fairness_fedfv, fednova, fedfa_huang2020fairness, fedadp}.

\begin{figure}[t]
     \centering
     \includegraphics[width=0.8\linewidth, trim=0 3.5cm 0 3cm, clip]{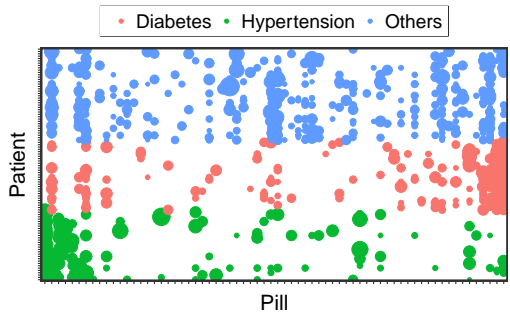}
     \caption{Distribution of pills collected from 100 real patients. Patients with identical diseases often take the same pills. Data can be classified into three groups: diabetes (red), hypertension (green), and others (blue).}
     \label{fig:pill_cluster}
 \end{figure}

However, the existing approaches suffer from two issues: incompleteness of the studied non-IID data types and non-generalization of the solution method. 
First, currently considered non-IID data types are limited and do not fully cover common patterns encountered in practice. 
Indeed, most previous approaches regard non-IID data as a context in which each client owns data samples with a set number of classes (\textit{label size imbalance})~\cite{fedavg_mcmahan2017communication,fedprox_li2020federated,fedat,ICPP2021_fedCav,ijcai2021_Fairness_fedfv,xiao2020averaging}
\cite{favor_hwangInforcom2020,fedadp}.
Recently, some works use \textit{label distribution imbalance} non-IID data to emulate different non-IID cases that are more close to the real-world data, e.g., distributing samples of a particular label to the clients following the power-law or Dirichlet distribution~\cite{li2019convergence,IWQOS2021_fedacs,fedfa_huang2020fairness,fednova}.
Unfortunately, all of the previous studies have only simulated the \emph{non-identical distribution} aspect, with no investigation into the \emph{non-independent} feature of the clients' data.
In reality, the data obtained from clients has a significant clustering feature, with many clients having similar labels. More particularly, certain clusters may have many clients, while others do not. For example, medical data from patients with the same disease will have a high degree of similarity, and common diseases will have many patients. Figure~\ref{fig:pill_cluster} illustrates the distribution of pills collected from $100$ real users. The users can be classified into three groups: diabetic patients, hypertensive patients, and the others. 
The data demonstrate an imbalance in two manners: 1) the data is disease-specific, i.e., patients with identical diseases often take the same medications, and 2) the data is statistically pill imbalance. There are numerous medications that are extensively used by all users, but there are also those that are not.
For more examples, see Section~\ref{sec:noniid}.
To fill in this gap, in this work, beside considering common types of non-IID data as existing studies (we name them as i.e., \textit{label skew non-IID}), we introduce a new non-IID data that exhibits \emph{non-independent} property of data across clients.  
Specifically, we focus on non-IID data having \textbf{inter-client correlation}, i.e., clients sharing a common feature can have correlated data. 
We consider that the global distribution of data labels is not uniform, and data labels frequently are partitioned into clusters\footnote{both overlap and non-overlap.}. That is, groups of clients owning the data classes belong to the same clusters. The number of clients per group is varied. We identify this scenario as \textit{cluster skew}. 

The second issue of the existing approaches,  non generalization of the solution method,  arises from the fact that the current methods are heuristic rules tailored explicitly for certain non-iid data types.
Consequently, they will only be applicable in a subset of non-iid circumstances. For example, in the case of our newly introduced \textit{cluster skew} scenario, the existing weight aggregation at the server may overly favor the DL models on the group with a larger number of clients even when the label size and number of samples of all the clients are balanced. 
To deal with the unfairness issue of the cluster skew non-IID data, we propose a novel aggregation method for \textbf{Fed}erated Learning with \textbf{D}eep \textbf{R}einforcement \textbf{L}earning, namely FedDRL.
The rationale behind our approach is that instead of designing explicit rules, we leverage the self-learning capability of the reinforcement learning agent to facilitate the server in discovering the optimum method to aggregate the clients' models. This way, the proposed approach does not rely on any specific data type. 
Specifically, the deep reinforcement learning model receives the state as the clients' local information (including the inference loss and the number of data samples), and then estimates the goodness of the action defined by the priorities assigned to the clients. 
Notably, to enhance the performance, we propose a novel two-stage training process that helps to enrich the training data and shorten the training time of our deep reinforcement learning model.
In summary, our main contributions in this work include:
\begin{itemize}
    \item We are the first to introduce and conduct a detail study on a novel non-iid data type named cluster-skew non-IID, which focuses on the inter-client correlation across the clients. 
    \item We then propose a novel method named FedDRL that explores Deep Reinforcement Learning in training a Federated learning model. To the best of our knowledge, this work is the first to use deep reinforcement learning to adaptively aggregate the clients' models. Along with the formulation of the deep learning framework used in the FL, we also propose a novel reward function and a two-stage training strategy that help to enhance the performance while reducing the time complexity. 
    \item We implement and measure the proposed FedDRL on three datasets. Intensive experiments show the FedDRL achieve better accuracy than the standard  FL methods including FedAvg~\cite{fedavg_mcmahan2017communication} and FedProx~\cite{fedprox_li2020federated}, e.g., up to $4.05$\% and $2.17$\% on average for CIFAR-100 dataset, respectively.
\end{itemize}
The remainder of the paper is organized as follows. Section~\ref{sec:background} provides a short summary of the standard federated learning and non-IID challenges. We then discuss our motivational study and present our proposal in Section~\ref{sec:method}. The experimental results are described in Section~\ref{sec:eval} and \ref{sec:discussion}. We finally conclude in Section~\ref{sec:conclusion}.


\section{Preliminaries} 
\label{sec:background}
\subsection{Federated Learning and FedAvg}
Federated Learning (FL) is the method that involves a large number of $K$ clients, usually hundreds to thousands, training locally and updating the Deep Learning (DL) model globally by using a weight aggregation method at a centralized server. The FL finds a DL model $w^*$ to minimize the objective function $f(w) = \sum^{K}_{k=1}\frac{n_k}{n}F_k(w)$, 
where $p_k = \frac{n_k}{n}$ denotes the relative sample size ($n = \sum^{K}_{k=1}n_k$), and $F_k(w) = \frac{1}{n_k}\sum_{i \in \mathcal{D}_k}l_i(w)$ is the local objective function of the client $i$. In which, $l_i(w)$ is the corresponding loss function for sample $i$ and $\mathcal{D}_k$ denotes the local dataset of the client $k$ ($n_k = |\mathcal{D}_k|$).


Federated Averaging (FedAvg)~\cite{fedavg_mcmahan2017communication} is the first and becomes one of the most commonly used FL methods to aggregate the locally trained models of the client into the shared global model. This method randomly selects a subset of $K$ clients in each communication round. Each client then performs the local training in $E$ epochs on its local dataset and uploads its local trained model to the server synchronously. Upon receiving local models from clients, the server the performs weighted aggregation as follows.
\begin{equation}
\label{equal:fedAvg}
\small
    n \gets \sum^{K}_{k=1}n_k; w^{t+1} \gets \sum_{k=1}^{K}{\frac{n_k}{n}w^{t}_k}.
\end{equation}
Such weight aggregation strategy  may lead to an unfair result where data distribution is imbalanced among clients (non-IID data)~\cite{fedfa_huang2020fairness}.

\subsection{Federated Learning with non-IID data}
\label{sec:noniid}
\subsubsection{Categories of non-IID Data}
In this work, We follow the taxonomy of non-IID (non independent and identical) data presented in ~\cite{icml2020_hsieh20a_skewscout} (Appendix-K) and~\cite{zhu2021federated}. Non-IID data can be formed by how data fail to be distributed identically across clients including:

\textbf{Feature skew or attribute skew}: clients have the same label sets\footnote{we use the terms \emph{labels} and \emph{classes} interchangeably in this work.} but features of samples are different between clients~\cite{zhu2021federated}. For example, in cancer diagnosis tasks, the cancer types are similar across hospitals but the radiology images can vary depending on the different imaging protocols or scanner generations used in hospitals~\cite{ICLR2021_li2021fedbn}.
    
\textbf{Quantity skew}: Different clients hold different amounts of data. For example, the younger generation may use mobile devices more frequently than the older generation, thus providing more local data to train the face recognition DL model~\cite{ijcai2021_Fairness_fedfv}.
    
\textbf{Label skew}: represents the scenarios where the label distribution differs between clients. In label size imbalance non-IID proposed by~\cite{fedavg_mcmahan2017communication}, each client has a fixed number of $c$ label classes; $c$ determines the degree of label imbalance. 
Recent works~\cite{li2019convergence,IWQOS2021_fedacs,fedfa_huang2020fairness,fednova} distribute the samples of a particular label to the clients following a particular distribution such as the power-law or Dirichlet distribution. Such kind of scenario is more close to the real-world data. For example, a car and a table can appear in both home cameras and streets cameras. However, cars commonly appear on streets camera while tables commonly appear on home cameras~\cite{luo2019real}.

This work focuses on label skew, which is common when data is generated from heterogeneous users. Unlike the previous research, we also consider data correlation between clients. Such kind of correlation has been observed in~\cite{icml2020_hsieh20a_skewscout} with their real-world dataset, i.e., Flickr-Mammal dataset~\cite{hsieh2020non}, and also can be observed from the number of cancer incidence cases in the USA from 2014 to 2018 for the cancer diagnosis tasks~\cite{cancer_review}.
In which clients who share the same feature, such as geographic region, timeline, or ages, may have the same data distribution.
First, global distribution of data label is not uniform, e.g, the most popular mammals (cat) has $23\times$ images greater than the less popular one (skunk)~\cite{icml2020_hsieh20a_skewscout}. Similarly, we found that number of incidence cases in the digestive system (the most popular) is $98\times$ of the incidence cases in eye and orbit (the less popular)~\cite{cancer_review}.
Second, we observe that data labels are frequently partitioned into clusters, and a group of clients owning the data labels belong to the same clusters. For example, users from Oceania usually share the image of kangaroos/koalas in Flick, while users from Africa share zebras/antelopes~\cite{icml2020_hsieh20a_skewscout}. Similarly, people under the age of 20 frequently get leukemia (28\%), while people between the ages of 20 to 34 frequently get cancer in the endocrine system (21\%)~\cite{cancer_review}.
In this paper, we name this phenomenon as \textbf{cluster skew}.

\subsubsection{Strategies for dealing with non-IID}
In this study, we focus on obtaining a DL model that (1) minimized  global objective function over the concatenation of all the local data and
(2) balanced the local objective functions of all the clients. Such target is also mentioned as the ``fairness" issue among clients~\cite{fedadp,ijcai2021_Fairness_fedfv}. There could be a different approach for each category of non-IID discussed above. 
Most of the existing work proposed to solve the statistical heterogeneity problems caused by non-IID dataset focused on the label skew and quantity skew non-IID~\cite{fedavg_mcmahan2017communication,fedprox_li2020federated,fedat,ICPP2021_fedCav,ijcai2021_Fairness_fedfv,xiao2020averaging,favor_hwangInforcom2020,fedadp, li2019convergence,IWQOS2021_fedacs,fedfa_huang2020fairness,fednova}. A few works target to the feature skew non-IID~\cite{ICLR2021_li2021fedbn}.
Kevin et. al.~\cite{icml2020_hsieh20a_skewscout} highlight the issue that we named cluster skew when building up their real-world dataset, yet focus on solving the communication overhead problem instead of the statistical heterogeneity problem.

\subsection{Basic of Reinforcement Learning}
\label{subsec:rl}
A Reinforcement Learning (RL) model is essentially comprised of four components: the \emph{agent}, the \emph{environment} with which the agent interacts and exploits, the \emph{reward} signal, which indicates how well the agent behaves regarding a specific target; and the \emph{policy}, which corresponds to the way the agent behaves. The ultimate goal of the agent is to maximize the total reward signal:
\begin{equation}
    \small
    \mathcal{G} = \sum_{t=1}^{T} \gamma^{t-1} r_t,
\end{equation}
where $\gamma$ is the discount factor and $r_t$ is the immediate reward signal at time step $t$. 

There are two main approaches to an RL problem, namely value-based learning and policy-based learning. 
Value-based methods focus on learning the value of states $V(s): \mathcal{s} \to \mathbb{R}$ or of state-action pairs $Q(s, a): \mathcal{S} \times \mathcal{A} \to \mathbb{R}$, where $\mathcal{S}$ and $\mathcal{A}$ are the state space and the action space, respectively. The state-value $V(s)$ depicts how good state $s$ is in terms of maximizing the total reward. The state-action value $Q(s,a)$ indicates how beneficial it is to perform action $a$ from a state $s$. Generally, when the agent receives the reward signal $r \in \mathbb{R}$, it performs a one-step bootstrap update using the well-known Bellman equation \cite{sutton2018reinforcement}. The optimal policy is deduced from the values learned by the agent adheres to the rationale. Once the values converge, the optimal action can be found by solving:
\begin{equation}
    a^* = \arg \max_a Q^*(s,a),  s\in \mathcal{S}, a \in \mathcal{A}(s).
\end{equation}
The most well-known algorithms are Q-learning and deep Q-learning (DQN \cite{sutton2018reinforcement}). \textcolor{black}{The expansion from a finite to an infinite number of states has motivated the adoption of a neural network} into reinforcement learning, which eventually gives rise to many states of the arts in recent years, including the policy-based methods.

Policy-based methods directly learn the optimal policy by continuously making adjustments upon the policy parameters in response to the reward signal received by the agent. In detail, policy-based methods, such as REINFORCE and Actor-Critic, maintain a $\theta$-parameterized neural network which takes state $s \in S$ as input and outputs the corresponding policy $\pi(s) = P(\cdot|s, \theta)$ describes the probability distribution over the action space at state $s$. The agent chooses an action accordingly and receives a reward signal $r$. Depending on the algorithm, a policy gradient theorem based update is applied upon the parameters $\theta$ with respect to a transformed value of the reward signal $r$, e.g., the temporal difference $\delta = r + \gamma V(s') - V(s)$.

\section{FedDRL: Federated learning with DRL-based adaptive aggregation}\label{sec:method}
\subsection{Motivation}
\begin{figure}
 \centering
    \includegraphics[width=0.9\linewidth]{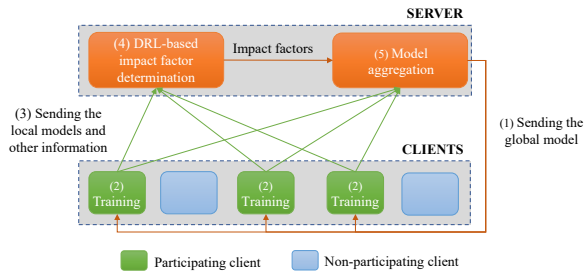}
     \caption{\label{fig:FedRL-overview}Workflow of FedDRL with 5 steps: (1) global model sending, (2) training at client, (3) local model sending, (4) impact factor determination, and (5) model aggregation.}
\end{figure}
In FL, the server plays a central role, receiving locally learned models (i.e., weights or gradients) from clients and aggregating them into a global model. Typically, the aggregation is performed by weighted summing of the clients' local models. Specifically, each client's model is assigned an impact factor representing its contribution to the global model, i.e., the higher the impact factor, the greater the influence on the global model. The impact factor assignment thus plays an essential role in deciding how well the server synthesizes the model.
The impact factors should be determined considering two criteria: 1) The goodness of knowledge acquired by the clients; 2) The correlation between the knowledge learned by the clients.
Regarding the former, it is clear that clients with the better knowledge should be given a higher impact factor.
Meanwhile, in terms of the latter, knowledge that repeatedly gained by multiple clients should be allocated a lower impact factor to avoid the model over fitting specific knowledge.
The most basic aggregation mechanism is proposed by FedAvg \cite{fedavg_mcmahan2017communication}, where impact factors are proportional to the number of samples possessed by the clients (Equation~(\ref{equal:fedAvg})). By assigning the importance of the clients proportional to the number of data samples, this approach considers all samples equally, and thus it is suitable exclusively for IID data.
Several recent methods determine impact factors heuristically based on the difference between clients' local model \cite{fedadp,fednova,fedfa_huang2020fairness}. In general, these strategies adhere to the rule of the majority; that is, knowledge acquired by a more significant number of clients is better. However, in practice, this is not always the case. Furthermore, this approach may lead to the general model being over-biased on data that occurs in many clients while disregarding unusual data.

Based on the above observations, we propose in this paper a novel model aggregation strategy based on Deep Reinforcement Learning (DRL). Instead of defining explicit rules for assigning impact factors, we leverage the capability of the reinforcement learning agent in interacting with the environment and self-learning to make the decision. This way, the proposed method can adapt to all kinds of client data distributions. This technique, in particular, can accommodate novel non-IID distributions that existing approaches have not investigated.

\subsection{Overview of FedDRL}
Figure~\ref{fig:FedRL-overview} illustrates the overview of our proposed DRL-based Federated learning model, FedDRL. 
The model is composed of a centralized server and several clients.
The clients utilize their local data to train models distributedly, while the server aggregates the models in a centralized manner.
Specifically, the server broadcasts the current global model to the clients at the beginning of each communication round.
After acquiring the global model, the clients use their data to train locally for a predefined number of epochs.
The clients then communicate the trained local models and other supporting information to the server.
The knowledge from the clients is first fed into a Deep reinforcement learning module, which determines the impact factor for each client's model.
With the impact factor defined, the server then aggregates the client models to produce a new global model.
\begin{figure*}
 \centering
 	\subfigure[Details of the DRL-based impact factor determination]{
 	 \centering
    \includegraphics[width=0.3\linewidth]{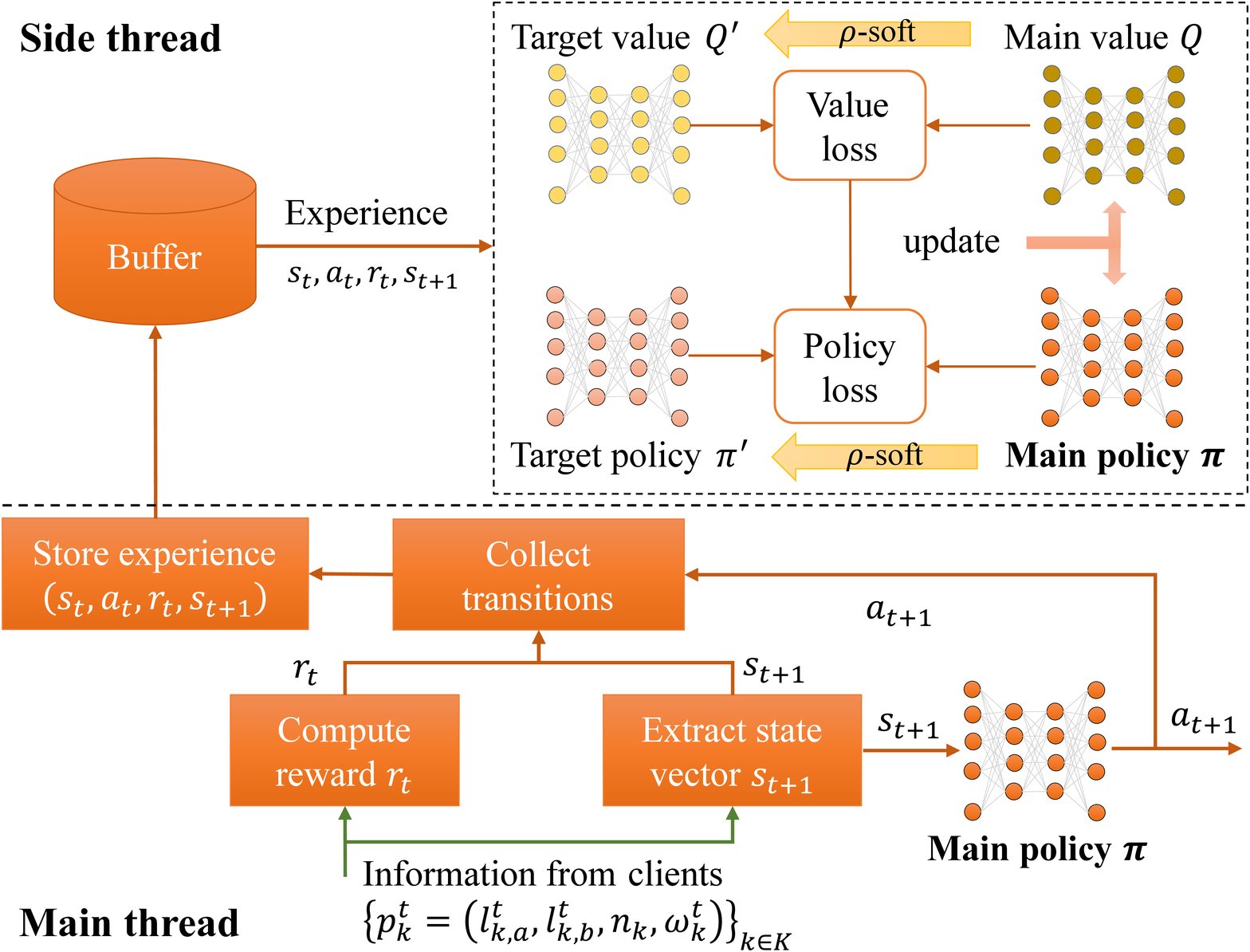}
    \label{fig:DRL-module}  	 }
 	 \hfill
 	\subfigure[The proposed two-stage training strategy]{
 	 \centering
    \includegraphics[width=0.33\linewidth]{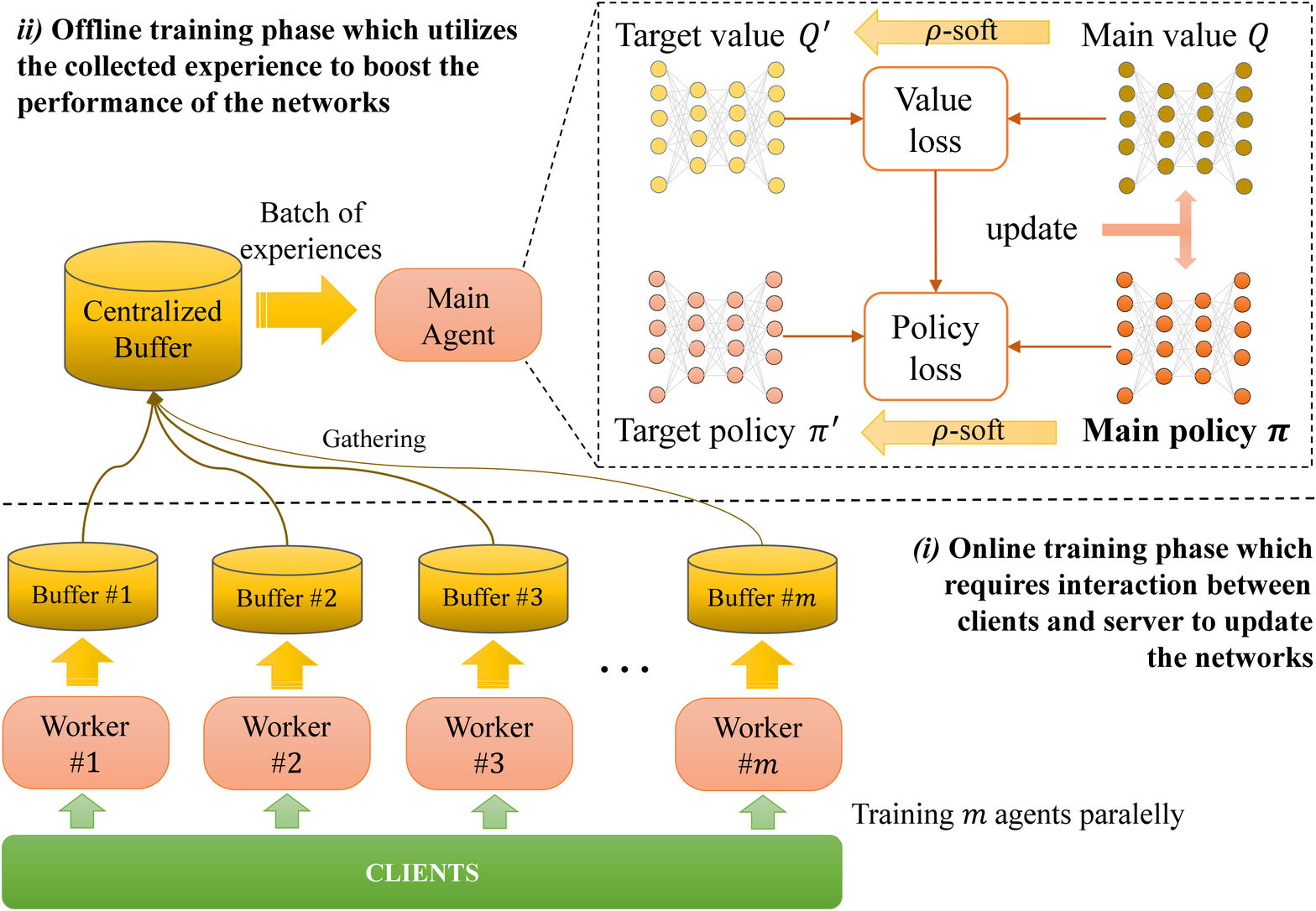}
    \label{fig:Two-stage} 
 	 }
 	 \hfill
 	\subfigure[Structures of the policy and value networks]{
 	 \centering
 	 \hspace{0.35cm}
    \includegraphics[width=0.21\linewidth]{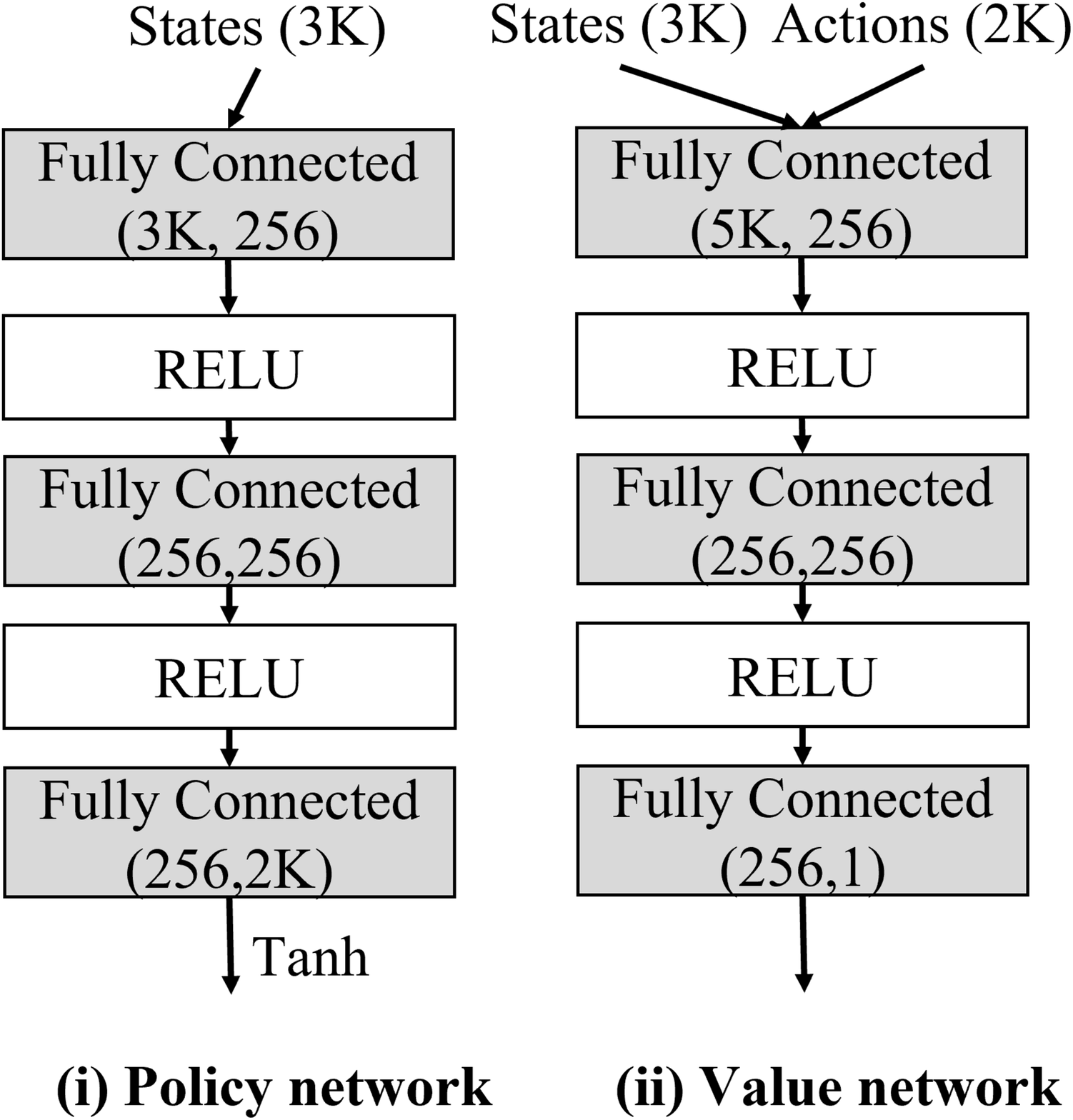}
     \hspace{0.35cm}
    \label{fig:network-structure} 
 	 }
 	 \caption{Details of the proposed FedDRL framework.}
 \end{figure*}

\subsection{Formulation of the Deep Reinforcement Learning Framework}
\label{subsec:fedDRL-framework}


We formulate the problem of model aggregation at the server of FL into a reinforcement learning paradigm. The following four elementary reinforcement learning concepts are defined.
\subsubsection{Agent}
We consider the situation where only one server is responsible for the aggregation and $N$ clients. For such, the server adopts a reinforcement learning based decision-making agent whose output are the impact factors indicating the contributions (in percentage) of locally trained models to the aggregated global one. Specifically, let $\vec{\alpha}^t$ and $\text{\textbf{W}}^t$ denote the impact factor vector and the concatenation of local models learned by the clients at round $t$, the server then computes the global model, e.g., usually the  gradients or weights of the DL model, as follows:
\begin{equation}
\small
\text{\textbf{w}}^{t+1}= \text{\textbf{W}}^t \cdot \vec{\alpha}^t = \sum_{k\in \mathcal{K}}\alpha_{k}^t \cdot w_{k}^{t},
\end{equation}
where $\mathcal{K}$ depicts the set of participating clients, and $\textbf{w}^{t}_{k}$ represents the $k$-th client's model at round $t$, $\textbf{w}^{t+1}$ denotes the global model at round $t+1$. Since we assume only a fraction of the clients participates in each round, the number of involved clients at every round, i.e., $|\mathcal{K}|$, is less than the total number of clients, i.e., $N$.

\subsubsection{State}
The state should provide any information that assists the decision-making process, i.e., determining the impact factors used to aggregate the clients' local models in our scenario.
The ultimate aim of the aggregation process at the server is to exploit meaningful information learned by the clients to create a global model that works well with all data. It implies that the state must contain information that allows the agent to determine: 1) whether knowledge gained by each client is valuable and 2) how this information should be aggregated to produce a global model that works effectively with all clients without bias toward any particular client.
As a result, the state should express the quality of models locally trained by clients and the relationship between the clients' data. 
To this end, we design the state to include the following information:
\begin{itemize}
    \item The losses of the global model on each client's datasets. These parameters are measured when the clients obtain the global model from the server (at the beginning of each communication round). It's worth noting that the models for all clients are identical at this time. Therefore, the losses reflect the correlation between the clients' data. At the same time, these losses also depicts how good the global model is for each client's data.
    \item The losses of local clients after performing local training at the end of each communication round. This parameter is calculated before the clients send their locally trained models to the server. This information tells us how good the a model trained by each client is.
    \item The amount of data held by each client. Clients with a different numbers of training data will, obviously, produce trained models with varying quality. Models trained on small datasets, in particular, are frequently less informative than models trained on large datasets. 
\end{itemize}
In summary, the state $s_t$ is a tuple of $3\times K$ parameters, $s_t = \{l_{1,b}^t, ..., l_{K,b}^t, $ $l_{1,a}^t, ..., l_{K,a}^t, n_1, ..., n_K \}$, where the first $K$ parameters are the losses evaluated right after the clients retrieve the global model from the server; $l_{1,a}^t, ..., l_{K,a}^t$ represent clients' losses evaluated when they finish the local training process at round $t$, and the finally $K$ parameters ${n_1, ..., n_K}$ depict the number of samples of the clients. 
\subsubsection{Action}
\label{subsec:FedRL:action}
The agent aims to identify the clients' impact factors, which are continuous numbers.
To this end, we define an action by a set of $K$ Gaussian distributions. 
Specifically, action $a_t$ is a tuple of $2 \times K$ values comprising the means and variances of $K$ Gaussian distributions, i.e., $a_t = \{ \mu_{1}^{t}, ..., \mu_{K}^{t}, \sigma_{1}^{t}, ..., \sigma_{K}^{t} \}$.
Given action $a_t$, the agent then determines the impact factor for the $k$-th client as follows:
\begin{equation}
    \small
    \alpha_{k}^{t} = \text{Softmax}(\mathcal{N}(\mu_{k}^{t}, \sigma_{k}^{t})).
\end{equation}
In addition, to enhance the model's stability throughout training, we add the following constraint:
\begin{equation}
    \small
    \vec{\sigma_t} \leq \beta \vec{\mu_t}, 
\end{equation}
where $\beta$ is a hyperparameter ranging from $0$ to $1$.

\subsubsection{Reward}
The design of the reward function is critical to the efficacy of the RL model since it directs the agent's behaviors.
We intend to create a global model that works well for all data, i.e., it should be suitable for all clients' data and not biased towards a specific client.
As a result, the reward function will have two objectives:
1) Improving the global model's accuracy across all clients.
2) Balancing the global model's performance over all clients' datasets.
We quantify the first goal by reducing the global model's average loss across all client datasets.
Besides, we achieve the second goal by minimizing the gap between the global model's maximum and minimum losses on against clients' datasets.
Accordingly, we design the reward as follows:
\begin{equation}
    \label{eq:reward}
    \small
    r_{t} = \left ( \frac{1}{K} \sum_{k=1}^{K} l_{k,b}^{t} \right ) + \left \{ \max_{k=1}^{K}(l_{k,b}^{t}) - \min_{k=1}^{K}(l_{k,b}^{t}) \right \},
\end{equation}
where the first term, i.e., $\frac{1}{K} \sum_{k=1}^{K} l_{k,b}^{t}$, represents the average losses of the global model against the local data of all clients, and the second term, i.e., $\max_{k=1}^{K}(l_{k,b}^{t}) - \min_{k=1}^{K}(l_{k,b}^{t})$, indicates the bias of the global model's performance on the clients' datasets. 

\subsection{Deep Reinforcement Learning Strategy}
\label{sec:DRL-training}
With the reinforcement learning terms defined in the previous section, in what follows, we describe the structure of our DRL network as well as the training strategies. 
We first present the structure of the model and a basic training strategy Section in \ref{sec:DRL-training-basic}.
To enhance the performance and reduce the training time, we then propose a two-stage training in Section \ref{sec:DRL-two-step-training}. 
\subsubsection{Basic training with temporal difference-based experience selection}
\label{sec:DRL-training-basic}
\begin{algorithm}[t]
\caption{\textbf{Basic training process}}
\label{algo:DDPG-update}
\small
\Input{$\theta$-parameterized main policy network $\pi$, $\phi$-parameterized main value network $Q$, experience buffer $\mathcal{D}$, target networks $\pi^{'}, Q^{'}$ as copies of $\pi$ and $Q$.}
For each experience $(s,a,r,s')$, assign a priority based on the temporal difference: $e$.prior $\gets |r + \gamma Q(s',a) - Q(s,a)|$;\\
Sort $\mathcal{D}$ by the descending order of the priority;\\
        \For{$b$ times updating}{
           Sample a batch $B = {(s,a,r,s')} \sim \mathcal{D}$;\\
            Compute targets:
            $y(r,s') \gets r + \gamma Q'(s', \pi'(s'))$;\\
            Update the main value network with gradient descent:
            $\nabla_\phi \frac{1}{|B|} \sum_{(s,a,r,s') \in B} (Q(s,a) - y(r,s'))^2$;\\
            Update the main policy network with gradient ascent:
            $\nabla_\theta \frac{1}{B} \sum_{s \in B} Q(s,\pi(s))$;\\
            Update target networks with:\\
            $\phi' \gets \rho \phi' + (1 - \rho)\phi$, $\theta' \gets \rho \theta' + (1 - \rho)\theta$;\\
        }
\end{algorithm}

Inspired by DDPG~\cite{lillicrap2015continuous}, we design a DRL agent as shown in Fig.~\ref{fig:DRL-module}.
The agent maintains two types of networks: the value network and the policy network.
The former is responsible for estimating the goodness of states, while the latter is for determining actions.
Moreover, for each type of network (i.e., value network or policy network), there are two subnets: the main network and the target network.
The main network and the target network have the same structure. The difference is that the main network directly interacts with the environment and is updated based on the feedback from the environment. Meanwhile, the target network is a soft copy of the main network; it is more stable and, therefore, is used as a reference point in training the main network.
The agent's operation is divided into two threads, running in parallel: the main thread and the side thread.
The former uses the main policy network to make decisions, i.e., calculating the impact factors of clients' models.
While the latter is responsible for improving the policy network's performance based on the feedback from the environment.
Specifically, the main thread works as follows.
\begin{itemize}
    \item At communication round $t+1$, the agent (i.e., server) receives from each $k$-th client a tuple $p_{k}^{t} = \{l_{k,b}^{t}, l_{k,a}^t, n_{k}, w_{k}^{t}\}$, where $l_{k,b}^{t}$, $l_{k,a}^{t}$ are the losses of the global model at client $k$, and the losses of the local model trained by client $k$ at round $t$, respectively, $n_k$ is the number of samples of client $k$ and $w_{k}^{t}$ is the weights of the model trained by client $k$. Agent then aggregates information received from all clients to establish state $s_{t+1}$.
    \item State $s_{t+1}$ will be fed into the main policy network \textbf{$\pi$} to derive action $a_{t+1}$. As mentioned in Section~\ref{subsec:FedRL:action}, action $a_{t+1}$ is a vector containing the means and variances of $K$ Gaussian distributions. The agent then calculates each client's impact factors and aggregates the clients' models. 
    \item Agent calculates reward $r_t$ using formula (\ref{eq:reward}) and stores a tuple of $(a_t, s_t, r_t, s_{t+1})$ into the experience buffer.
\end{itemize}

Parallelly with the main thread, the agent also runs the side thread continuously to improve the policy network's performance.
In the following, we will describe the details of our basic training process used in the side thread.
The pseudo-code of our basic training method is shown in Algorithm~\ref{algo:DDPG-update}.

\begin{table}[tb]
    \centering
    \caption{Configuration of the policy and value networks}
    \small
    \begin{tabular}{|m{5cm}|m{2cm}|}
        \hline
         \textit{Hyper-parameters} & \textit{Values}\\
         \hline
         $\pi$-network's \#layer & 3\\
         $Q$-network's \#layer & 3\\
         Hidden layer size & 256 \\
         $\pi$-network learning rate & 0.0001 \\
         $Q$-network learning rate & 0.001 \\
         Experience buffer size & 100000 \\
         Discount factor $\gamma$ & 0.99 \\
         Soft main-target update factor $\rho$ & 0.02 \\
         \hline
    \end{tabular}
    \label{tab:DDPQ-configuration}
    \vspace{-0.1cm}
\end{table}

\begin{itemize}
    \item Firstly, the agent extracts from the experience buffer a set of experiences. Each experience is a tuple of $(a_t, s_t, r_t, s_{t+1})$. Here, we apply the temporal difference-prioritizing strategy to prioritize the experiences. Specifically, each experience $e_t = (a_t, s_t, r_t, s_{t+1})$ will be assigned a priority $\delta_t = r_t + \gamma  Q(s_{t+1},a_t)-Q(s_t,a_t)$. The higher the priority, the likely the experience is chosen (Lines 1-2 in Algorithm \ref{algo:DDPG-update}).
    \item Using the selected experiences, the agent will update the main value network using the gradient descent method, and update the main policy network using the gradient ascent method (Lines 4-7 in Algorithm \ref{algo:DDPG-update}).
    \item Finally, updates of the main network will be transferred to the target network using $\rho$-soft update mechanism (Line 8-9 in Algorithm \ref{algo:DDPG-update}).
\end{itemize}
Figure \ref{fig:network-structure} illustrates the structure of the policy and value networks. 
Specifically, we construct our policy networks with $3$ fully connected layers, each contains $256$ nodes, in paired with Leaky Relu activation function. 
The output of a policy network is a flatten vector of $K \times 2$ representing the mean and standard deviation of the multi-variate Gaussian distribution from which we sample the impact factor. For the value networks, we use $2$ hidden layers of size $256$, the outputs of which are also activated by a Leaky Relu function. 
The full configuration is represented in Table \ref{tab:DDPQ-configuration}.
\begin{algorithm}[t]
\renewcommand*{\algorithmcfname}{Algorithm}
\caption{\textbf{Overall flow of FedDRL}}
\label{algo:FedRL}
\small
\Input{A server with an $w$-parameterized DNN, $N$ clients with separated datasets, maximum communication round $T$, $K$ participating clients at each round.}
    \textbf{Server:} Initiate $\theta$-parameterized policy network $\pi$, $\phi$-parameterized value network $Q$, experience buffer $\mathcal{D}$, target networks $\pi^{'}, Q^{'}$ as copies of $\pi$ and $Q$;\\

    \textbf{Server:} Initiate global model $w^{0}$;\\
    \For{communication round $t$ from $0$ to $T$}{
        Server sends $K$ copies of $w^{t}$ to the participating clients;\\
        \For{each client $k$ in $K$ clients \textbf{in parallel}}{
            $w_{k}^{(t,0)} \gets w^{t}$  \\
            $l_{k,b}^t \gets $ Inference loss with $w_{k}^{(t,0)}$\\
            Local training in $E$ epochs with batch $b$:\\
                \hskip2.5em 
                $w^{t}_k \gets w^{t} - \eta.\frac{1}{b}\sum_{e=1}^{E}\sum_{i = 1}^{\frac{n_k}{b}}\nabla l_i(w_{k}^{(t,i)})$ \\
            $l_{k,a}^{t} \gets $ Inference loss with $w_{k}^{t}$   \\ 
            Send $p_{k}^{t} \gets $($l_{k,b}^{t}$, $l_{k,a}^{t}$, $n_k$, $w^{t}_k$) to the server;\\
        }
        Server unpacks $p_{k}^{t}$ from clients, concatenates to form state $s_{t+1}$;\\
        Server selects action: $(\vec{\mu}, \vec{\sigma}) \gets \pi(s) + \epsilon, \epsilon \sim \mathcal{N}$;\\
        Server computes impact factor vector: $\vec{\alpha} \gets \text{softmax}(N(\vec{\mu}, \vec{\sigma}))$;\\ 
        Server performs weighted aggregation upon concatenated parameters-matrix from clients: $\text{\textbf{w}}^{t+1}= \text{\textbf{W}}^t \cdot \vec{\alpha}^t = \sum_{k\in \mathcal{K}}\alpha_{k}^t \cdot w_{k}^{t}$;\\
        Server observes the next state $s'$, computes reward $r$;\\
        Server stores $(s,a,r,s')$ into $\mathcal{D}$ ;\\
        \If{$\mathcal{D}$ is sufficient}{
            Server performs \textbf{procedure in Algorithm~\ref{algo:DDPG-update}};\\
        }
}
\end{algorithm}

\subsubsection{Two-stage training strategy}
\label{sec:DRL-two-step-training}
In the basic training strategy mentioned above, each communication round provides only one experience. As a consequence, the number of experiences utilized to train the policy and value networks is relatively small. This constraint will limit the performance of the networks as a consequence. 
To this end, we propose a novel two-stage training technique as shown in Fig~\ref{fig:Two-stage}.
In the first stage, i.e., named online training phase, we create $m$ identical agents (which we name workers). 
These agents simultaneously interact with the environment to train the networks and generate experiences. 
It is worth noting that although the workers are initially identical, they will evolve into distinct individuals over the training process, and hence their collected experiences will differ.
The workers' experiences are then merged to build an experience buffer, which will be utilized to train a main agent in the second stage. The main agent is trained in an offline manner without interaction with the environment.
Specifically, we employ the experience from the experience buffer created by the first stage to update the main agent's policy and value networks using the gradient ascent and gradient descent methods, respectively. With such a training strategy, we dramatically enrich the training data while decreasing the training time.
The trained main agent is then used to make the decision.

We summarize the total flow of FedDRL in Algorithm \ref{algo:FedRL}.


\subsection{Assumptions and Limitations}
The assumptions and limitations of the FedDRL are as follows:

\textbf{Targeted communication strategies}: Our method is designed for synchronous FL where the server wait for all the clients finished their local training (with a fixed training epochs\footnote{Unless otherwise stated, we fix all the number of epochs of all the clients to $5$.}) before starting to compute the new global model. The technique in this work may not be suitable for supporting the asynchronous FL~\cite{fedat} where the global weight is updated right after receiving the locally trained model from one (or some) clients (over $K$ client involved each round). Note, however, that our technique is still applicable to other communication techniques such as using spare data compression~\cite{2020_Sattler, icdcs2020_sparcification} or hierarchical architecture~\cite{ijcai2021_HFL}.

\textbf{Overhead of FedDRL}: Although FedDRL requires some extra computation and communication overhead, it is still practical in real-world applications. Computation overhead is the inference latency at the end of each training round that is trivial when the size of local datasets of clients is usually small. It is reported that such overhead is less than one second in MNIST~\cite{ICPP2021_fedCav}.
It is worthy to noting that performing the aggregation at the server also requires trivial overhead for calculating the impact factors, i.e., the inference phase through the policy network of DRL module. We further estimate this overhead in Section~\ref{sec:overhead}.
For communication overhead, our FedDRL only needs some extra floating points numbers for the inference loss in comparison with the FedAvg.

\section{Experiments}\label{sec:eval}
\subsection{Evaluation Methodology}
This section describes a wide range of experiments to compare the proposed FedDRL against two baselines including FedAvg~\cite{fedavg_mcmahan2017communication} and FedProx\cite{fedprox_li2020federated}. We evaluate the classification performance using the best top-1 accuracy on test sets of the targeted dataset.
We also evaluate the SingleSet for being references, i.e., training all the data samples of all the clients in a single machine (e.g., training at the server or in the system of only one client)
\footnote{SingleSet can be the best when the total number of samples across all the clients is big enough. In this case, the training with SingleSet becomes similar to the training on an IID dataset.}.

\subsubsection{Federated Datasets}
We evaluate the accuracy of a deep neural network (DNN) model trained on a non-IID partitioned dataset. In this work, we pick up datasets and DNN models that are curated from prior work in federated learning. In detail, we use three different federated datasets, i.e., CIFAR-100~\cite{Krizhevsky09}, Fashion-MNIST~\cite{xiao2017fashion}, and  MNIST~\cite{Mnist}. 
To study the efficiency of different federated methods, it is necessary to specify how to distribute the data to the clients and simulate data heterogeneity scenarios. We study three ways to partition the targeted datasets by considering the label size imbalance, quantity skew, and clustered skew (more detail in Section~\ref{sec:background}). Although our main target in this evaluation is the dataset distributed with the cluster-skew non-IID, we also estimate the performance of our proposal in a conventional way, e.g., Pareto. We distribute the training samples among \textit{N} clients, i.e., 10 and 100 clients, as follows:
\begin{itemize}
    \item Pareto (denoted as \textbf{PA}): non-IID partition method where the number of samples of a label among clients following a power law~\cite{li2019convergence, fedprox_li2020federated}. We assign $2$ labels to each client for the MNIST dataset (20 labels/client for the CIFAR-100). 
    \item Clustered-Equal (denoted as \textbf{CE}): 
    We consider the simple case of cluster skew. We first arrange the clients into groups with a group that has a significantly higher number than the others, e.g., the main group with $\delta\times N$ clients. The higher $\delta$ means more bias toward the main group. We also fix the number of labels in each client to two. The number of samples per client does not change among clients.
    \item Clustered-Non-Equal (denoted as \textbf{CN}): 
    Similar to CE but the number of samples per client is unbalanced.
\end{itemize}

\begin{table}[bt]
	\caption{Characteristic of non-IID partitioned methods. $\checkmark$: related non-IID type; $\times$: not related. Explanation in remarks.} 
	\label{table:noniid}
	\centering
	\scriptsize
	\setlength\tabcolsep{3pt} 
	\resizebox{0.48\textwidth}{!}{%
		\begin{tabular}{lcccl}
			\toprule
			\textbf{\begin{tabular}[c]{@{}c@{}}Partition \\ Method\end{tabular}} & 
			\textbf{\begin{tabular}[c]{@{}c@{}}Clustered \\ Skew\end{tabular}} &
			\textbf{\begin{tabular}[c]{@{}c@{}}Label Size \\ Imbalance\end{tabular}} &
			\textbf{\begin{tabular}[c]{@{}c@{}}Quantity \\ Imbalance\end{tabular}} &
			\textbf{Remarks}
			\\ \midrule 
			
			
			
			PA & $\times$ & $\checkmark$ & $\checkmark$ &  \begin{tabular}[l]{@{}c@{}} $\#$samples 
			follows a \\power law~~\cite{li2019convergence}\end{tabular} \\ \midrule
			
			CE & $\checkmark$ & $\checkmark$ & $\times$ &  Our proposed method \\ \midrule
			
			CN & $\checkmark$ & $\checkmark$ & $\checkmark$ & Our proposed method\\
			

			\bottomrule
		\end{tabular}
	}
\end{table}
 \begin{figure}
     \centering
     \small
     \subfigure[PA]{
        \label{fig:PA}
        \includegraphics[width=0.3\linewidth]{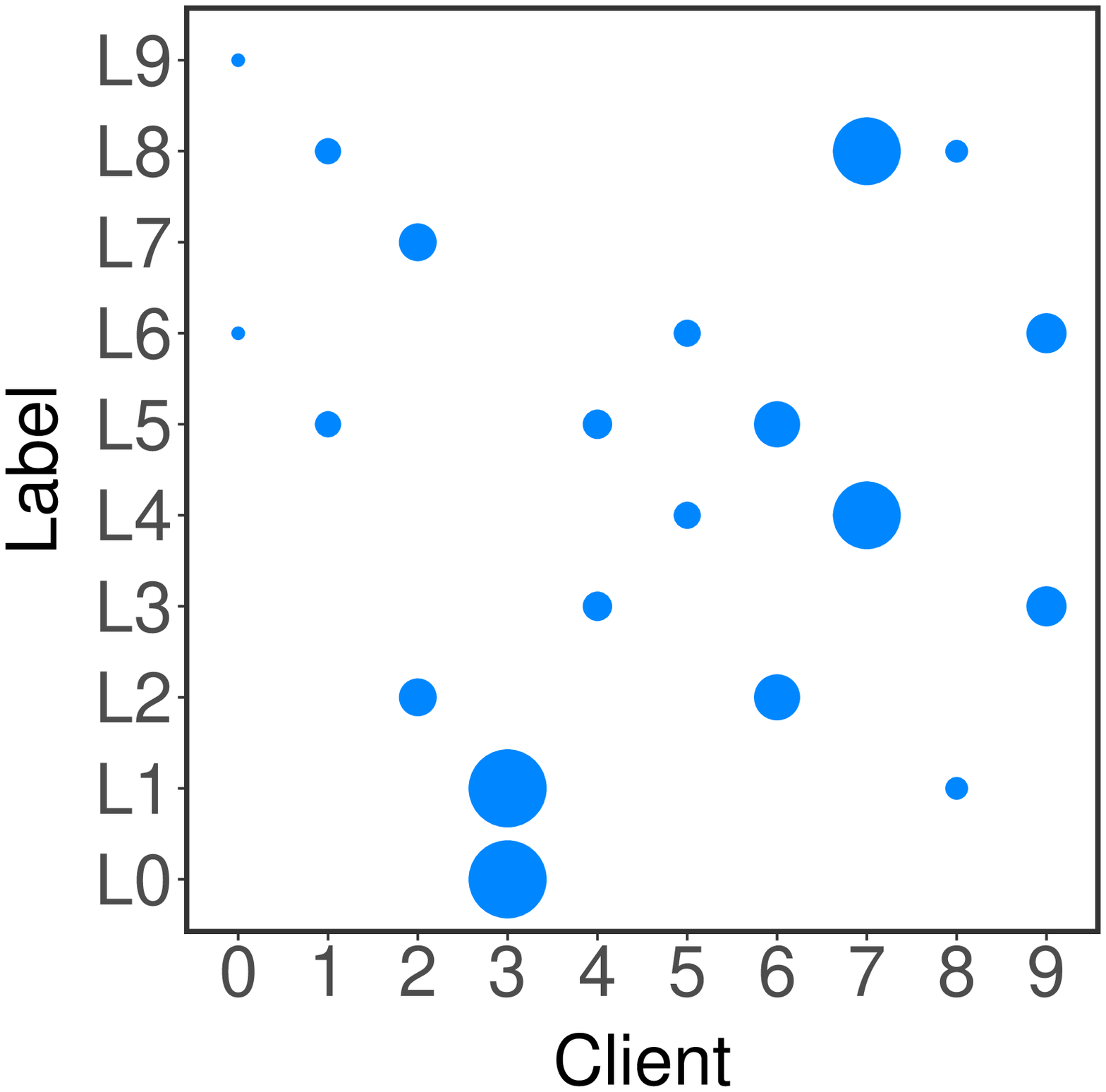}
     }
      \subfigure[CE]{
        \label{fig:CE}
        \includegraphics[width=0.3\linewidth]{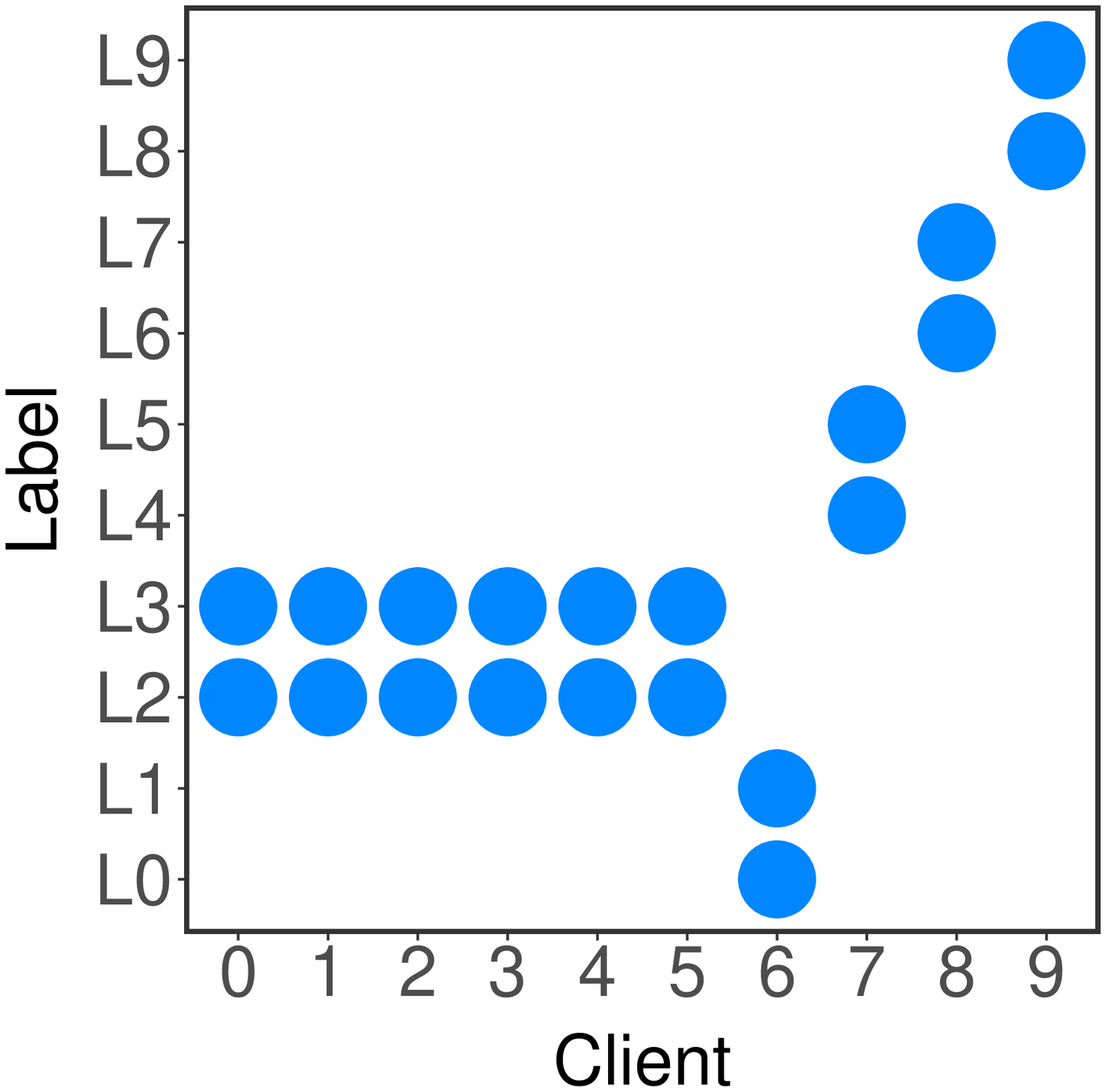}
     }
      \subfigure[CN]{
        \label{fig:CN}
        \includegraphics[width=0.3\linewidth]{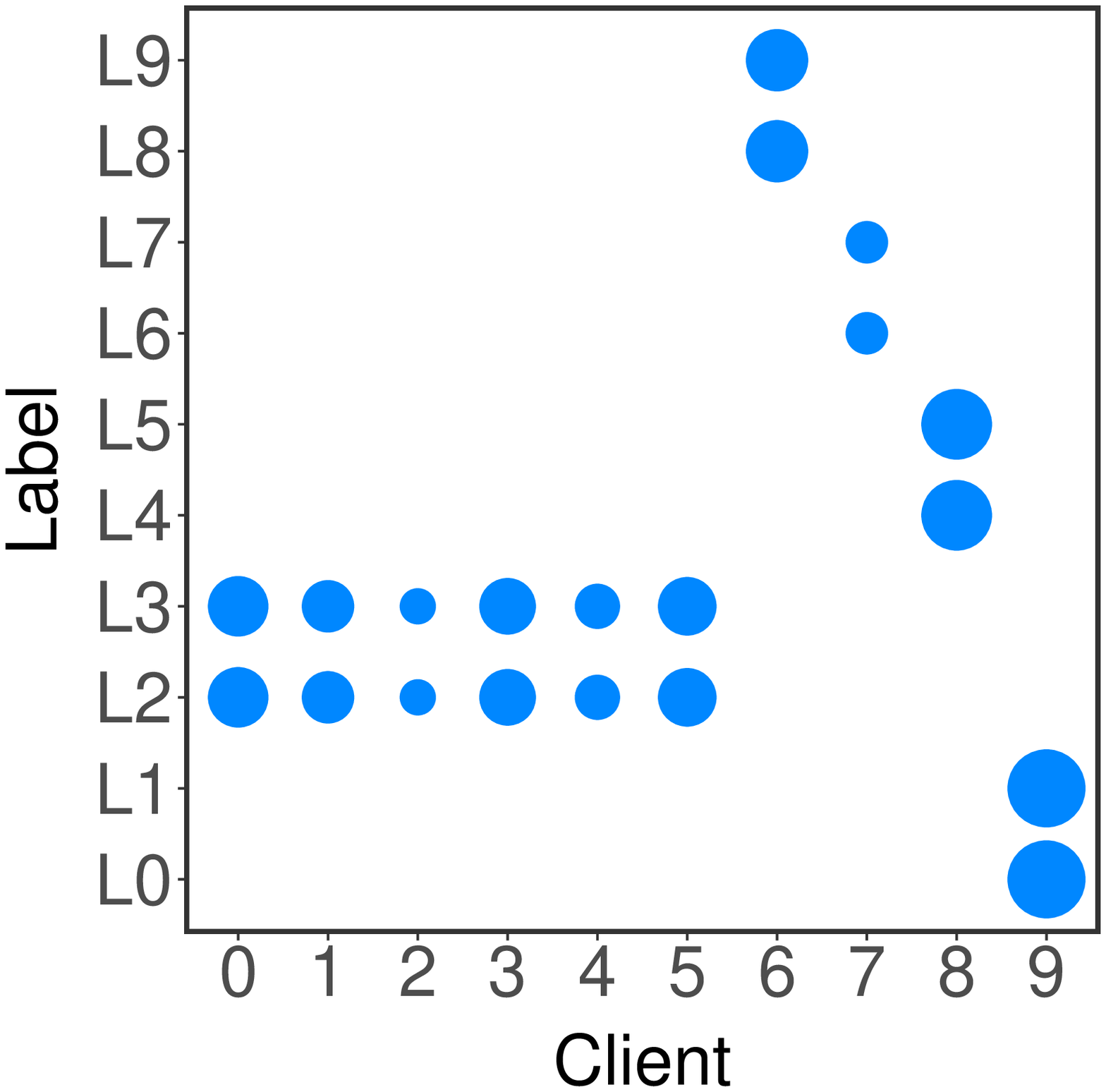}
     }
     \caption{Illustration of data partitioning methods with 10 clients. The size of the circle shows the number of samples/client.}
     \label{fig:dataset}
 \end{figure}

The characteristic of the targeted partition method is summarized in Table~\ref{table:noniid}. For example, Figure~\ref{fig:dataset} illustrates how the dataset is distributed to clients with different partitioning methods in the case of 10 clients.


\subsubsection{Network architecture and training procedure}
We train a simple convolutional neural network (CNN) on MNIST and Fashion-MNIST datasets as mentioned in~\cite{fedadp}. For CIFAR-100 dataset, we use VGG-11 network~\cite{SimonyanZ14a} as it is in \cite{2020_Sattler,fednova}. We use the de-factor solver in DL, i.e., SGD (stochastic gradient descent), as the local solver with the local epochs $E=5$, a learning rate of $0.01$ and a local batch size $b=10$ for all the experiments. 
We evaluate with the system of 10 and 100 clients like the prior work in federated learning~\cite{fedavg_mcmahan2017communication, fedprox_li2020federated, fedat}.
If not specified, our default setting for the number of participating clients at each communication round is $K=10$ . We train all the FL methods in $1,000$ communication rounds. 
For the FedProx method, we set the proximal term $\mu=0.01$. For all the experiments, we simulate FL on a computer that consists of 2 Intel Xeon Gold CPUs, and 4 NVIDIA V100 GPUs. We deploy only one experiment on an entire GPU.

\subsubsection{Configuration of our deep reinforcement learning model}
We implemented/trained FedDRL in Pytorch version 1.8.1.
The details of the policy and value networks are as described in Table ~\ref{tab:DDPQ-configuration}.
Concerning the two-stage training strategy, we set the number of workers to $2$ for computational reasons.


\begin{table}[t]
	\caption{\small
	Comparison of top-1 test accuracy to baseline approaches with different partitioning methods, i.e., PA, CE and CN. 
	The values show the best accuracy that each FL method reaches during training. The best performance results are highlighted in the bold, underlined font. \emph{impr.(a)} and \emph{impr.(b)} are the relative accuracy improvement of FedDRL compared with the best and the worst baseline FL method (in percentage), respectively.  Significant improvements, i.e., $>1\%$, are also marked in the bold, underlined font.} 
	\label{table:accuracy}
	\centering
	\scriptsize
	\setlength\tabcolsep{2pt} 
	\resizebox{\linewidth}{!}{%
    \begin{tabular}{@{}cl|lll|lll|lll@{}}
    \toprule
    \multicolumn{2}{l}{}
    & \multicolumn{3}{c}{\textbf{CIFAR-100}}
    & \multicolumn{3}{c}{\textbf{Fashion-MNIST}}
    & \multicolumn{3}{c}{\textbf{MNIST}}             \\
    \cmidrule(lr){3-5}\cmidrule(lr){6-8}\cmidrule(lr){9-11}
    \multicolumn{2}{c}{\begin{tabular}[c]{@{}c@{}}Partitioning \\method\end{tabular}}
    &  \multicolumn{1}{c}{PA} & \multicolumn{1}{c}{CE} & \multicolumn{1}{c}{CN}     
    &  \multicolumn{1}{c}{PA} & \multicolumn{1}{c}{CE} & \multicolumn{1}{c}{CN}
    &  \multicolumn{1}{c}{PA} & \multicolumn{1}{c}{CE} & \multicolumn{1}{c}{CN} \\
    \midrule
    \multirow{5}{*}{\begin{tabular}[c]{@{}c@{}}10\\  clients\end{tabular}}  
    & SingeSet 
    & 75.63 & 68.6 & 72.67          
    & 90.06 & 88.9 & 88.48 
    & 97.65 & 98.87 & 99.01 \\
    \cmidrule(lr){3-11}
    & FedAvg   
    & 69.81          & 62.18          & 64.6          
    & 86.19 & 82.49 & 82.97 
    & 89.58 & 98.13 & 97.98 \\
    & FedProx  
    & 71.13 & 62.25          & 66.8           
    & 86.54 & 82.52 & 82.43 
    & 97.23 & 96.89 & 96.88 \\
    & FedDRL   
    & \textbf{\underline{72.63}}           & \textbf{\underline{64.51}} & \textbf{\underline{67.32}} 
    & \textbf{\underline{86.92}} & \textbf{\underline{82.59}} & \textbf{\underline{84.26}} 
    & \textbf{\underline{97.31}} & \textbf{\underline{98.13}} & \textbf{\underline{98.08}} \\
    \cmidrule(lr){3-11}
    & \emph{impr.(a)} 
    & \textbf{\underline{2.11\%}}\% & \textbf{\underline{3.63\%}} & 0.78\%   
    & 0.44\% & 0.08\% & \textbf{\underline{1.55\%}} 
    & 0.08\% & 0.00\% & 0.10\% \\
    & \emph{impr.(b)} 
    & \textbf{\underline{4.18\%}} & \textbf{\underline{3.75\%}} & \textbf{\underline{4.21\%}} 
    & 0.85\% & 0.12\% & \textbf{\underline{2.22\%}}
    & \textbf{\underline{8.63\%}} & \textbf{\underline{1.28\%}} & \textbf{\underline{1.24\%}} \\
    \midrule
    \multirow{5}{*}{\begin{tabular}[c]{@{}c@{}}100 \\ clients\end{tabular}} 
    & SingeSet 
    & 64.55  & 71.26 & 68.73
    & 88.9 & 88.60 & 88.35 
    & 99.19 & 98.79 & 98.82 \\
    \cmidrule(lr){3-11}
    & FedAvg   
    & 58.29  & 60.95 & 58.17
    & 85.57 & 84.88 & 83.82 
    & 98.43 & 97.9 & 98.22 \\
    & FedProx  
    & 56.03  & 55.55 & 60.53
    & 86.36  & 84.78 & 83.89 
    & 98.20 & 97.95 & 98.13 \\
    & FedDRL   
    & \textbf{\underline{58.41}}  & \textbf{\underline{61.12}}& \textbf{\underline{61.51}}
    & \textbf{\underline{86.65}} &\textbf{\underline{86.71}} & \textbf{\underline{86.69}} 
    & \textbf{\underline{98.63}} & \textbf{\underline{98.03}} & \textbf{\underline{98.27}} \\
    \cmidrule(lr){3-11}
    & \emph{impr.(a)}    
    & 0.21\% & 0.28\% & \textbf{\underline{1.62\%}}
    & 0.34\% & \textbf{\underline{2.16\%}} & \textbf{\underline{3.34\%}} 
    & 0.20\% & 0.08\% & 0.05\%\\
    & \emph{impr.(b)} 
    & \textbf{\underline{4.25\%}} & \textbf{\underline{10.03\%}} & \textbf{\underline{5.74\%}}
    & \textbf{\underline{1.26\%}} & \textbf{\underline{2.28\%}} & \textbf{\underline{3.42\%}} 
    & 0.44\% & 0.13\% & 0.14\% \\
    \bottomrule
    \end{tabular}
    }
\end{table}


\subsection{Experimental Results}
\subsubsection{Classification accuracy}
\begin{figure*}
     \centering
     \small
        \label{fig:cifar_accuracy}
        \includegraphics[width=1\linewidth,trim=0 0cm 0 1cm]{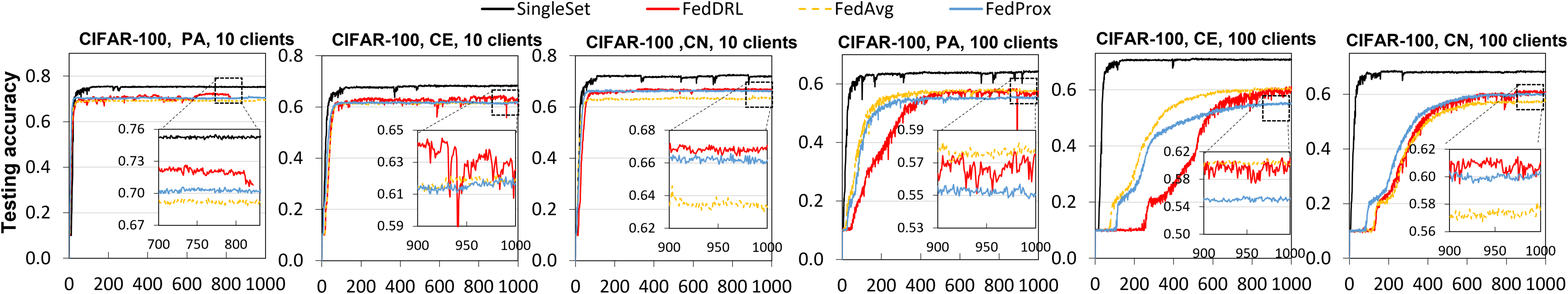}\\
        \includegraphics[width=1\linewidth,trim=0 0.5cm 0 0cm]{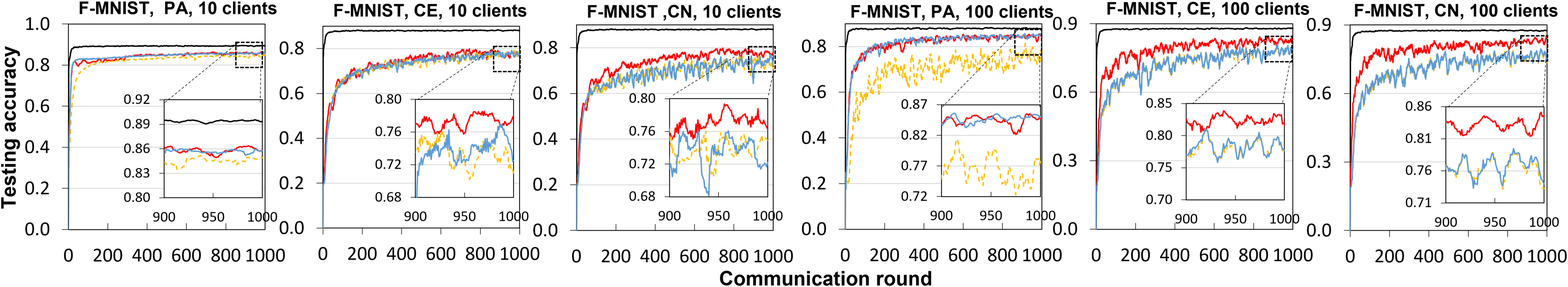}\\
     \caption{Top-1 test accuracy of different FL methods vs. communication round. The results of Fashion-MNIST are
     are plotted with the average-smoothed of every $10$ communication round to have a better visualization. We omit the result of MNIST due to the space limitation.}
     \label{fig:accuracy}
 \end{figure*}

Table~\ref{table:accuracy} presents the results of the classification accuracy when comparing our proposed FedDRL to the baseline methods on all the datasets and partitioning methods. We report the best top-1 test accuracy that each FL method reaches within 1000 communication rounds. For the CE and CN partitioning methods, we set the non-IID level $\delta=0.6$ in this evaluation. 
Specifically, FedDRL achieves better accuracy than all other two baseline methods. The accuracy difference can also be noticed from the top-1 accuracy timeline graphs shown in Figure~\ref{fig:accuracy}. Especially, when the number of clients becomes high, e.g., 100 clients versus 10 clients, the improvement of FedDRL on accuracy becomes larger.

For the CIFAR-100 dataset with a small number of clients, e.g., 10 clients, FedDRL outperforms the best baseline FL method, FedProx, by $2.17$\% and the worst baseline FL method, FedAvg, by $4.05$\% on average. Those are $0.7$\% and $6.67$\% when the number of clients becomes as large as $100$, respectively. Especially with the CE and CN data partitioning methods, our target in this work, the accuracy of FedAvg and FedProx are dropped significantly (more than 10\%) when there exists a bias of the client-correlation. 
In those cases, FedDRL can significantly improve the baselines.
For example, when 100 clients are involved in the network, FedDRL improves the accuracy up to 10,03\%. This result emphasizes that our weighted aggregation method using Reinforcement learning can be more effectively engage the clients from the `smaller' groups\footnote{Groups with a smaller number of clients}, leading to better accuracy.

For the MNIST and Fashion-MNIST datasets, FedDRL is still slightly better than the baseline methods. This is an expected result because the image classification tasks in those cases are simple with only $10$ different classes. 
Thus, the accuracy of all the methods is asymptotic to the accuracy of the SingleSet, which leads to no room for optimizing. 
As the result, the gaps between the test accuracy of FedDRL and the best baseline FL method are trivial, i.e., around 1\% . Thus, we consider FedDRL equivalent to FedAvg and FedProx when training on MNIST and Fashion-MNIST.
\begin{figure}[t]
     \centering
     \includegraphics[width=1\linewidth,trim=0 0.5cm 0 1.5cm]{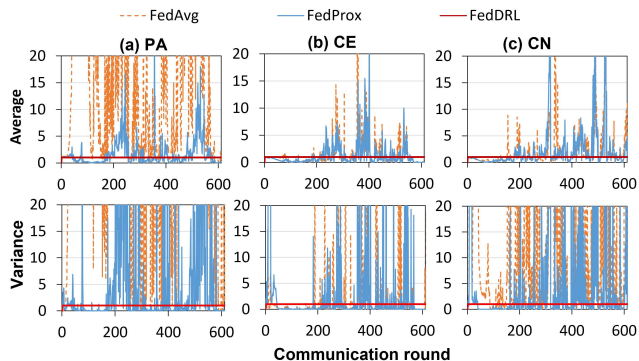}
     \caption{Average (top row) and variance (bottom row) of inference loss among all clients, normalized to those of FedDRL (red line) in the cases of CIFAR-100 dataset, 10 clients.}
     \label{fig:inference_loss}
 \end{figure}

\subsubsection{Robustness to the client datasets}
Testing accuracy on a test dataset that is usually independent and identical helps estimate the goodness of a Deep Learning (DL) model over the global distribution. However, it is reported that the average accuracy may be high but the accuracy of each device may not always be high~\cite{fedfa_huang2020fairness} when the server overly favors some clients during its weight aggregation (due to the bias of the non-IID dataset). 
To estimate the robustness of an FL method against clients, we test the global model on the local sub-dataset of all the participating clients at the beginning of each communication round, i.e., do the inference pass at clients.
Figure~\ref{fig:inference_loss} shows the average (in the top row) and variance (in the bottom row) of the inference loss among all the clients. Values are normalized to those of FedDRL (the red line). A value of a FL method that are under the red line refer that it achieves a lower average inference loss (or variances) and vise versa.

Figure~\ref{fig:inference_loss} shows that FedDRL has consistently lower inference loss on clients than the baseline across all experiments.
FedAvg and FedProx observe significantly higher inference loss. This is due to the effect of the impact factor assigned for each client when the server performs the weight aggregation. In which, the clients belonging to the main groups will contribute less to the 
global model than the client belonging to a `small' group. The result also implies that the inference loss of FedDRL at some beginning communication rounds are much higher than those of the baselines because this is the time when the DRL module learns how to assign the impact factor promptly. After 200-300 communication rounds, the inference loss at clients of FedDRL becomes smaller than those of the baseline.

\subsection{Sensitivity Analysis}
\begin{figure*}
	\centering
	\begin{minipage}[!t]{0.45\textwidth}
     \includegraphics[width=1\columnwidth, trim=0 0.5cm 0 1cm]{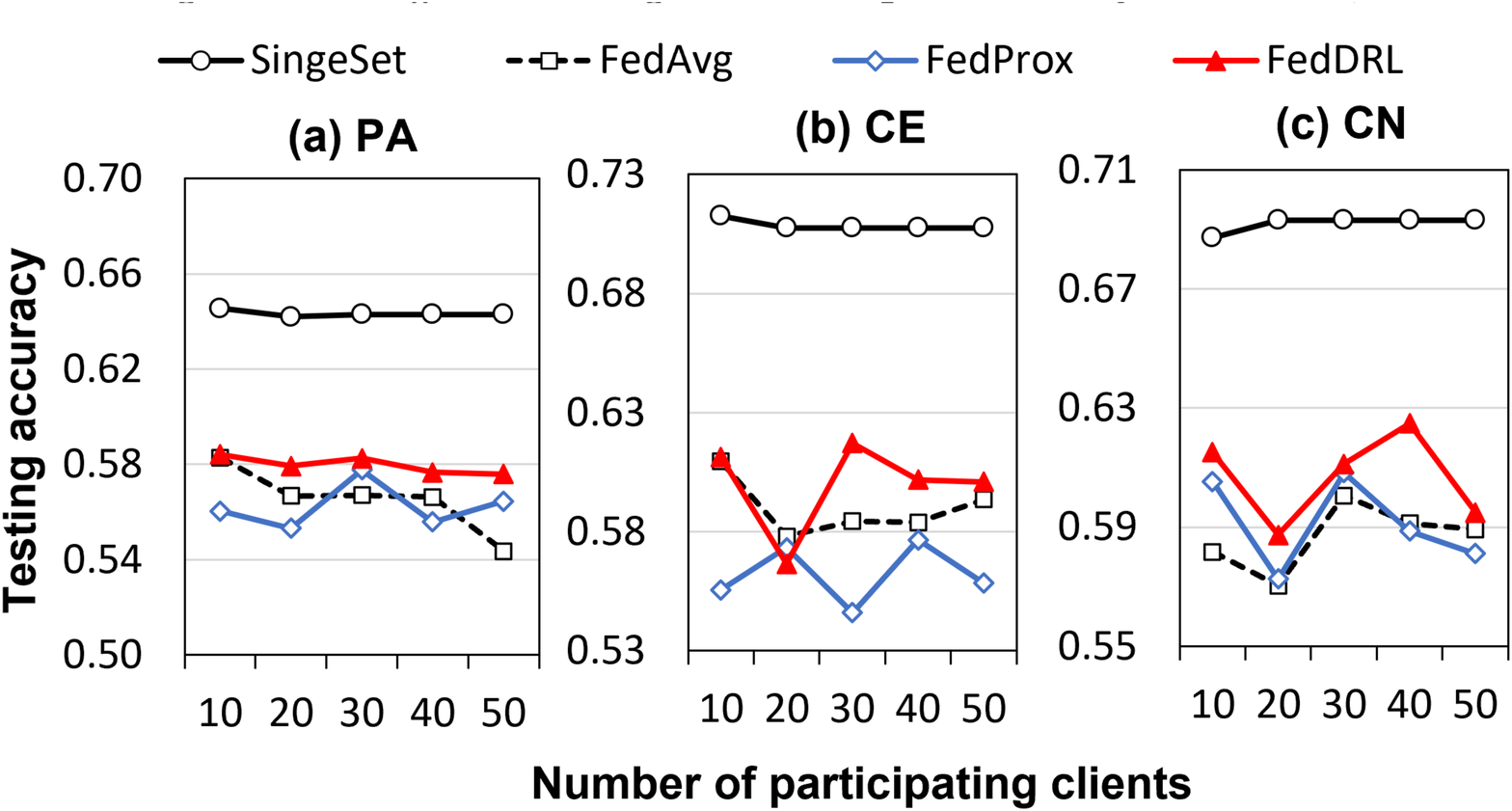}
     \caption{Testing accuracy vs. the number of participating clients $K$ (CIFAR-100, $N = 100$ clients).}
     \label{fig:cifar_participating}
 	\end{minipage}
	\hspace{4pt}
	\begin{minipage}{0.31\textwidth}
     \includegraphics[width=1\linewidth,trim=0 0.5cm 0 1cm]{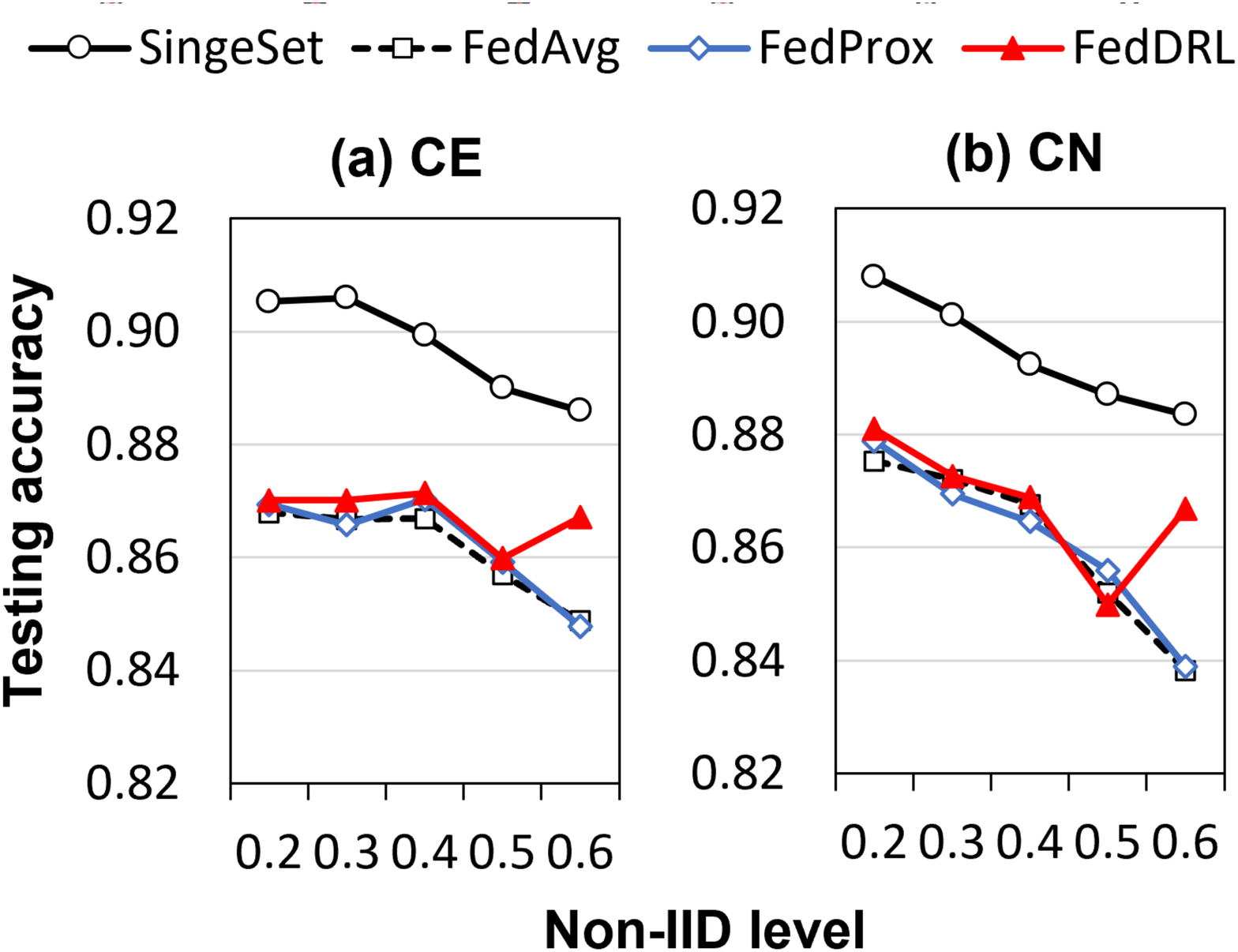}
     \caption{Testing accuracy vs. the non-IID level (Fashion-MNIST, 100 clients).}
     \label{fig:fmnist_level}
 	\end{minipage}
 	\hspace{4pt}
 	\begin{minipage}{0.20\textwidth}
     \includegraphics[width=1\linewidth,trim=0 0.5cm 0 1cm]{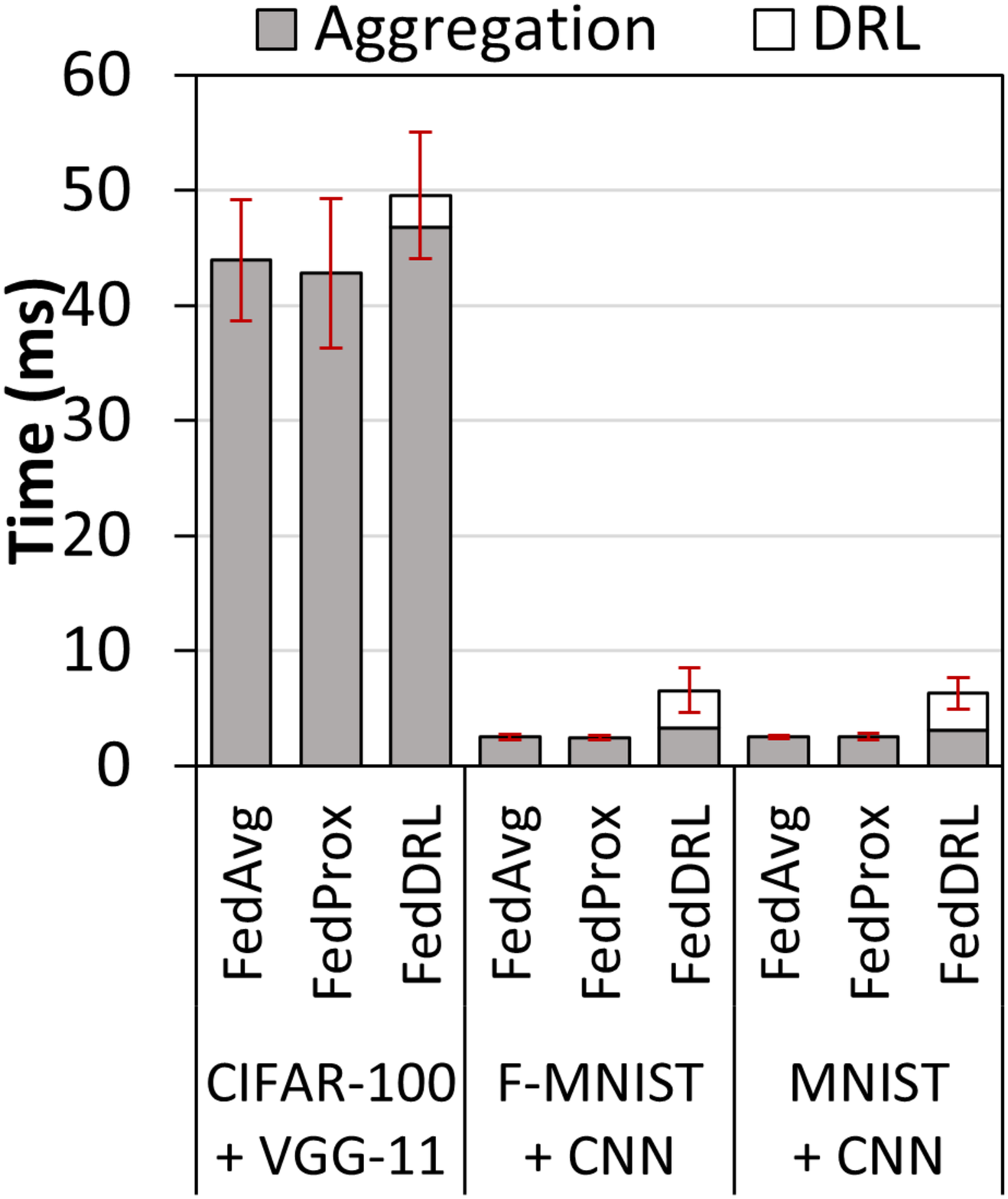}
     \caption{Average server computation time.}
     \label{fig:compute_overhead}
 	\end{minipage}
  \end{figure*}
\subsubsection{Impact of the number of participating clients}
We next conduct a sensitivity study to quantify the impact of the client participation level on accuracy. As shown in Figure~\ref{fig:cifar_participating}, 
we change the number of participating clients $K$ at one communication round from $10$ up to $50$ over the total $N = 100$ clients when training on the CIFAR-100 dataset.
We observe that varying the number of participating clients would affect the convergence rate but would not impact the accuracy eventually when the training converges.
As the result, the improvement in accuracy of FedDRL over the other two baseline methods is consistently maintained.

\subsubsection{Impact of the non-IID level}
It is well known that the convergence behaviors of FL methods are sensitive to the degree of non-IID. Our target in this work is to study the effect of the cluster-skew on testing accuracy. We thus change the number of clients belonging to the main group ($\delta$) to change the level of bias caused by the clients from the main group. 
Figure~\ref{fig:fmnist_level} show how the testing accuracy changed when $\delta$ varied from 0.2 to 0.6 in the case of Fashion-MNIST dataset. 
First, increasing the non-IID level negatively affects the testing accuracy for all the FL methods. 
Second, the results also indicate that FedDRL mitigates the negative impact of these biases. 
For instance, FedDRL is slightly better than the baselines in most cases. Especially, when $\delta=0.6$, there are significant improvements in the accuracy of FedDRL over FedAvg and FedProx, e.g., approximately $2-3\%$. 

\section{Discussion}
\label{sec:discussion}
\subsection{Impact of the non-IID type}

\begin{table}[t]
	\caption{
	\revision{Top-1 test accuracy with label size imbalanced non-iid suggested by~\cite{fedavg_mcmahan2017communication} on the CIFAR-100 dataset.}}
	\label{table:accuracy2}
	\centering
	\scriptsize
	\setlength\tabcolsep{1pt} 
	\resizebox{\linewidth}{!}{%
    \begin{tabular}{@{}l|cccc|cccc@{}}
    \toprule
    \multirow{2}{*}{\begin{tabular}[c]{@{}c@{}}\# \\clients\end{tabular}}
    & \multicolumn{4}{c}{\textbf{Equal}}
    & \multicolumn{4}{c}{\textbf{Non-equal}}             \\
    \cmidrule(lr){2-5}\cmidrule(lr){6-9}
    &  SingleSet & FedAvg & FedProx     
    &  \multicolumn{1}{c}{FedDRL} &  SingleSet & FedAvg & FedProx     
    &  FedDRL \\
    \midrule
    10 & 81.16 & 75.52 & 70.18 & \textbf{\underline{76.65}} & 81.56 & 75.5 & 76.65 & \textbf{\underline{76.9}}\\
    100 & 81.34 & 72.71 & 72.52 & \textbf{\underline{73.47}} & 81.59 & 72.44 & 72.89 & \textbf{\underline{73.42}}\\
    \bottomrule
    \end{tabular}
    }
\end{table}
In this section, we illustrate that our approach does not rely on any particular data type. It is applicable not just to cluster-skew but to label-skew in general.  Specifically, we consider two common label size imbalanced non-IID suggested by ~\cite{fedavg_mcmahan2017communication} as follow:
\begin{itemize}
    \item \textbf{E}qual: non-IID partition method provided by FedAvg~\cite{fedavg_mcmahan2017communication}. The dataset is sorted and separated into $2N$ shards, where $N$ is the number of clients. Each client has samples of two shards, so that the total number of samples per client does not change among clients. 
    \item \textbf{N}on-equal: non-IID partition method provided by FedAvg~\cite{fedavg_mcmahan2017communication}. The dataset is sorted and separated into $10N$ shards. Each client has samples of a random number of shards ranged from $6$ to $14$. 
\end{itemize}
In Section~\ref{sec:eval}, we showed numerical data regarding the Pareto distribution, in which FedDRL demonstrated a significant increase in accuracy compared to other benchmarks. A similar observation was observed on the equal and non-equal data distribution. 
FedDRL, for example, outperforms the best baseline FL algorithms by 1.85\% to 2.13\% on the CIFAR100 dataset (as shown in Table~\ref{table:accuracy2}.)

\subsection{Convergence rate}
\begin{figure*}
     \centering
     \includegraphics[width=1\linewidth,trim=0 0.5cm 0 1cm, clip]{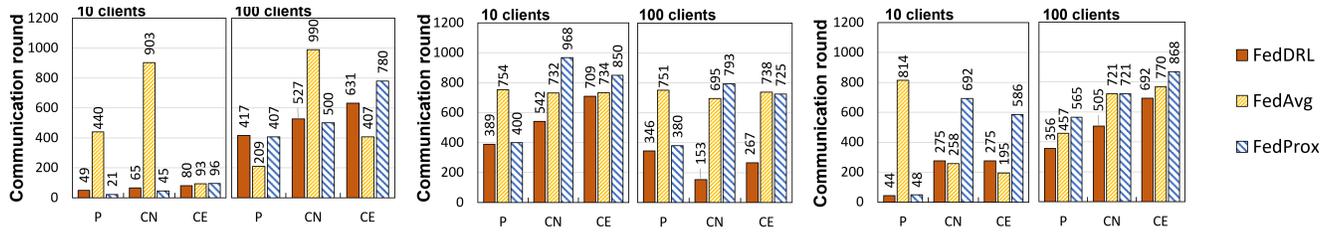}
     \vspace{-0.2cm}
     \caption{Convergence rate of different FL methods on CIFAR-100 , Fashion-MNIST, and MNIST datasets.}
     \label{fig:coverage}
 \end{figure*} 
Figure~\ref{fig:coverage} presents the comparison of the number of communication rounds it takes for each FL method to achieve a target testing accuracy. In this evaluation, we select the target testing accuracy as the minimum accuracy over FedDRL and the two baselines (as shown in Table~\ref{table:accuracy}). For example, to reach an accuracy of $62,1$\% for the CIFAR-100 dataset with the CE partitioning method in a network of $10$ clients, FedDRL requires $80$ communication rounds. FedAvg and FedProx spend $1.16\times$ and $1.2\times$ longer than FedDRL, respectively. 
Overall, the convergence rate of FedDRL is not always the fastest. It is interesting to emphasize that the convergence rates of FedAvg and FedProx do not remain consistent across different data partitioning methods.  Instead, our method FedDRL always converges as fast as the fastest one in all the cases. The results on Fashion-MNIST and MNIST show a similar trend. This result shows that FedRDL is much more consistent and robust than the baseline methods. It is due to the usage of adaptive priorities for clients based on their local data in the proposed method.
\subsection{Computational overhead of FedDRL}
\label{sec:overhead}
As mentioned, our FedDRL is designed for synchronous federated learning, yet significant computation time on the server would prolong the synchronization latency.
To ensure that our weighted aggregation method using DRL does not contribute much computation overhead at the server versus other baseline FL methods, we estimate the time to calculate the impact factors (DRL in Figure~\ref{fig:compute_overhead}). We also present the time to perform the actual weights aggregation for reference (Aggregation).
The result shows that the computation time of the DRL module is trivial, e.g., only $~3$ milliseconds. Such overhead is consistent with all the models and datasets. By contrast, the aggregation time of FedDRL depends on the size of Deep Learning models., e.g., $~45$ and $3$ milliseconds for VGG-11 and CNN, respectively.
In short, our FedDRL is practical in terms of computation time at the server.\\[-0.5cm] 


\section{Conclusion}\label{sec:conclusion}
Despite the popularity of the standard federated learning (FL) FedAvg~\cite{fedavg_mcmahan2017communication} and its alternatives, there are some critical challenges when applying it to the real-world non-IID data. 
Most existing methods conduct their studies on very restricted non-IID data, e.g., label-skew and quantity-skew non-IID.
In this work, we tackled a novel non-IID type observed in real-world datasets, where groups of clients exhibit a strong correlation and have similar datasets.  We named it the cluster-skew non-IID. 
To deal with such kind of non-IID data, we then proposed a  novel  aggregation  method for Federated  Learning, namely  FedDRL, which exploits Deep Reinforcement Learning approach. 
We performed intensive experiments on three common datasets, e.g., CIFAR-100, Fashion-MNIST, and MNIST, with three different ways to partition the datasets. The experiment result showed that the proposed FedDRL achieves better accuracy than FedAvg~\cite{fedavg_mcmahan2017communication} and FedProx~\cite{fedprox_li2020federated}, e.g., absolutely up to $10\%$ in accuracy.

\section*{Acknowledgements}
This work was funded by Vingroup Joint Stock Company (Vingroup JSC), Vingroup, and supported by Vingroup Innovation Foundation (VINIF) under project code VINIF.2021.DA00128. 

\bibliographystyle{ACM-Reference-Format}
\bibliography{refer_full}


\begin{thebibliography}{31}


\ifx \showCODEN    \undefined \def \showCODEN     #1{\unskip}     \fi
\ifx \showDOI      \undefined \def \showDOI       #1{#1}\fi
\ifx \showISBNx    \undefined \def \showISBNx     #1{\unskip}     \fi
\ifx \showISBNxiii \undefined \def \showISBNxiii  #1{\unskip}     \fi
\ifx \showISSN     \undefined \def \showISSN      #1{\unskip}     \fi
\ifx \showLCCN     \undefined \def \showLCCN      #1{\unskip}     \fi
\ifx \shownote     \undefined \def \shownote      #1{#1}          \fi
\ifx \showarticletitle \undefined \def \showarticletitle #1{#1}   \fi
\ifx \showURL      \undefined \def \showURL       {\relax}        \fi
\providecommand\bibfield[2]{#2}
\providecommand\bibinfo[2]{#2}
\providecommand\natexlab[1]{#1}
\providecommand\showeprint[2][]{arXiv:#2}

\bibitem[Ben{-}Nun and Hoefler(2018)]%
        {ben2018demystifying}
\bibfield{author}{\bibinfo{person}{Tal Ben{-}Nun} {and}
  \bibinfo{person}{Torsten Hoefler}.} \bibinfo{year}{2018}\natexlab{}.
\newblock \showarticletitle{Demystifying Parallel and Distributed Deep
  Learning: An In-Depth Concurrency Analysis}.
\newblock \bibinfo{journal}{\emph{CoRR}}  \bibinfo{volume}{abs/1802.09941}
  (\bibinfo{year}{2018}).
\newblock
\showeprint[arxiv]{1802.09941}


\bibitem[Chai et~al\mbox{.}(2021)]%
        {fedat}
\bibfield{author}{\bibinfo{person}{Zheng Chai}, \bibinfo{person}{Yujing Chen},
  \bibinfo{person}{Ali Anwar}, \bibinfo{person}{Liang Zhao},
  \bibinfo{person}{Yue Cheng}, {and} \bibinfo{person}{Huzefa Rangwala}.}
  \bibinfo{year}{2021}\natexlab{}.
\newblock \showarticletitle{{FedAT: A High-Performance and
  Communication-Efficient Federated Learning System with Asynchronous Tiers}}.
  In \bibinfo{booktitle}{\emph{Proceedings of the International Conference for
  High Performance Computing, Networking, Storage and Analysis}}
  \emph{(\bibinfo{series}{SC '21})}. Article \bibinfo{articleno}{60},
  \bibinfo{numpages}{16}~pages.
\newblock


\bibitem[Cho et~al\mbox{.}(2020)]%
        {cho2020client}
\bibfield{author}{\bibinfo{person}{Yae~Jee Cho}, \bibinfo{person}{Jianyu Wang},
  {and} \bibinfo{person}{Gauri Joshi}.} \bibinfo{year}{2020}\natexlab{}.
\newblock \showarticletitle{Client selection in federated learning: Convergence
  analysis and power-of-choice selection strategies}.
\newblock \bibinfo{journal}{\emph{arXiv preprint arXiv:2010.01243}}
  (\bibinfo{year}{2020}).
\newblock


\bibitem[Han et~al\mbox{.}(2020)]%
        {icdcs2020_sparcification}
\bibfield{author}{\bibinfo{person}{Pengchao Han}, \bibinfo{person}{Shiqiang
  Wang}, {and} \bibinfo{person}{Kin~K. Leung}.}
  \bibinfo{year}{2020}\natexlab{}.
\newblock \showarticletitle{Adaptive Gradient Sparsification for Efficient
  Federated Learning: An Online Learning Approach}. In
  \bibinfo{booktitle}{\emph{2020 IEEE 40th International Conference on
  Distributed Computing Systems (ICDCS)}}. \bibinfo{pages}{300--310}.
\newblock


\bibitem[{Howlader N, Noone AM, Krapcho M, et al. (eds), National Cancer
  Institute}(2021)]%
        {cancer_review}
\bibfield{author}{\bibinfo{person}{{Howlader N, Noone AM, Krapcho M, et al.
  (eds), National Cancer Institute}}.} \bibinfo{year}{2021}\natexlab{}.
\newblock \bibinfo{title}{{SEER Cancer Statistics Review 1975-2018}}.
\newblock
  \bibinfo{howpublished}{\url{https://seer.cancer.gov/csr/1975_2018/browse_csr.php}}.
\newblock
\newblock
\shownote{[19 April 2021]}.


\bibitem[Hsieh et~al\mbox{.}(2020a)]%
        {icml2020_hsieh20a_skewscout}
\bibfield{author}{\bibinfo{person}{Kevin Hsieh}, \bibinfo{person}{Amar
  Phanishayee}, \bibinfo{person}{Onur Mutlu}, {and} \bibinfo{person}{Phillip
  Gibbons}.} \bibinfo{year}{2020}\natexlab{a}.
\newblock \showarticletitle{The Non-{IID} Data Quagmire of Decentralized
  Machine Learning}. In \bibinfo{booktitle}{\emph{Proceedings of the 37th
  International Conference on Machine Learning}}
  \emph{(\bibinfo{series}{Proceedings of Machine Learning Research},
  Vol.~\bibinfo{volume}{119})}. \bibinfo{publisher}{PMLR},
  \bibinfo{pages}{4387--4398}.
\newblock


\bibitem[Hsieh et~al\mbox{.}(2020b)]%
        {hsieh2020non}
\bibfield{author}{\bibinfo{person}{Kevin Hsieh}, \bibinfo{person}{Amar
  Phanishayee}, \bibinfo{person}{Onur Mutlu}, {and} \bibinfo{person}{Phillip
  Gibbons}.} \bibinfo{year}{2020}\natexlab{b}.
\newblock \showarticletitle{The non-iid data quagmire of decentralized machine
  learning}. In \bibinfo{booktitle}{\emph{International Conference on Machine
  Learning}}. PMLR, \bibinfo{pages}{4387--4398}.
\newblock


\bibitem[Huang et~al\mbox{.}(2020)]%
        {fedfa_huang2020fairness}
\bibfield{author}{\bibinfo{person}{Wei Huang}, \bibinfo{person}{Tianrui Li},
  \bibinfo{person}{Dexian Wang}, \bibinfo{person}{Shengdong Du}, {and}
  \bibinfo{person}{Junbo Zhang}.} \bibinfo{year}{2020}\natexlab{}.
\newblock \showarticletitle{Fairness and accuracy in federated learning}.
\newblock \bibinfo{journal}{\emph{arXiv preprint arXiv:2012.10069}}
  (\bibinfo{year}{2020}).
\newblock


\bibitem[Krizhevsky and Hinton(2009)]%
        {Krizhevsky09}
\bibfield{author}{\bibinfo{person}{A. Krizhevsky} {and} \bibinfo{person}{G.
  Hinton}.} \bibinfo{year}{2009}\natexlab{}.
\newblock \showarticletitle{Learning multiple layers of features from tiny
  images}.
\newblock \bibinfo{journal}{\emph{Master's thesis, Dep.\ of Comp.\ Sci.\,
  Univ.\ of Toronto}} (\bibinfo{year}{2009}).
\newblock


\bibitem[Lecun et~al\mbox{.}(1998)]%
        {Mnist}
\bibfield{author}{\bibinfo{person}{Y. Lecun}, \bibinfo{person}{L. Bottou},
  \bibinfo{person}{Y. Bengio}, {and} \bibinfo{person}{P. Haffner}.}
  \bibinfo{year}{1998}\natexlab{}.
\newblock \showarticletitle{Gradient-based learning applied to document
  recognition}.
\newblock \bibinfo{journal}{\emph{Proc. IEEE}} \bibinfo{volume}{86},
  \bibinfo{number}{11} (\bibinfo{year}{1998}), \bibinfo{pages}{2278--2324}.
\newblock
\urldef\tempurl%
\url{https://doi.org/10.1109/5.726791}
\showDOI{\tempurl}


\bibitem[Li et~al\mbox{.}(2020b)]%
        {li2020federated}
\bibfield{author}{\bibinfo{person}{Tian Li}, \bibinfo{person}{Anit~Kumar Sahu},
  \bibinfo{person}{Ameet Talwalkar}, {and} \bibinfo{person}{Virginia Smith}.}
  \bibinfo{year}{2020}\natexlab{b}.
\newblock \showarticletitle{Federated learning: Challenges, methods, and future
  directions}.
\newblock \bibinfo{journal}{\emph{IEEE Signal Processing Magazine}}
  \bibinfo{volume}{37}, \bibinfo{number}{3} (\bibinfo{year}{2020}),
  \bibinfo{pages}{50--60}.
\newblock


\bibitem[Li et~al\mbox{.}(2020c)]%
        {fedprox_li2020federated}
\bibfield{author}{\bibinfo{person}{Tian Li}, \bibinfo{person}{Anit~Kumar Sahu},
  \bibinfo{person}{Manzil Zaheer}, \bibinfo{person}{Maziar Sanjabi},
  \bibinfo{person}{Ameet Talwalkar}, {and} \bibinfo{person}{Virginia Smith}.}
  \bibinfo{year}{2020}\natexlab{c}.
\newblock \showarticletitle{Federated optimization in heterogeneous networks}.
\newblock \bibinfo{journal}{\emph{Proceedings of Machine Learning and Systems}}
   \bibinfo{volume}{2} (\bibinfo{year}{2020}), \bibinfo{pages}{429--450}.
\newblock


\bibitem[Li et~al\mbox{.}(2020a)]%
        {li2019convergence}
\bibfield{author}{\bibinfo{person}{Xiang Li}, \bibinfo{person}{Kaixuan Huang},
  \bibinfo{person}{Wenhao Yang}, \bibinfo{person}{Shusen Wang}, {and}
  \bibinfo{person}{Zhihua Zhang}.} \bibinfo{year}{2020}\natexlab{a}.
\newblock \showarticletitle{{On the Convergence of FedAvg on Non-IID Data}}. In
  \bibinfo{booktitle}{\emph{8th International Conference on Learning
  Representations, {ICLR} 2020, Addis Ababa, Ethiopia, April 26-30, 2020}}.
\newblock


\bibitem[Li et~al\mbox{.}(2021)]%
        {ICLR2021_li2021fedbn}
\bibfield{author}{\bibinfo{person}{Xiaoxiao Li}, \bibinfo{person}{Meirui
  JIANG}, \bibinfo{person}{Xiaofei Zhang}, \bibinfo{person}{Michael Kamp},
  {and} \bibinfo{person}{Qi Dou}.} \bibinfo{year}{2021}\natexlab{}.
\newblock \showarticletitle{Fed{BN}: Federated Learning on Non-{IID} Features
  via Local Batch Normalization}. In \bibinfo{booktitle}{\emph{International
  Conference on Learning Representations}}.
\newblock


\bibitem[Lillicrap et~al\mbox{.}(2015)]%
        {lillicrap2015continuous}
\bibfield{author}{\bibinfo{person}{Timothy~P Lillicrap},
  \bibinfo{person}{Jonathan~J Hunt}, \bibinfo{person}{Alexander Pritzel},
  \bibinfo{person}{Nicolas Heess}, \bibinfo{person}{Tom Erez},
  \bibinfo{person}{Yuval Tassa}, \bibinfo{person}{David Silver}, {and}
  \bibinfo{person}{Daan Wierstra}.} \bibinfo{year}{2015}\natexlab{}.
\newblock \showarticletitle{Continuous control with deep reinforcement
  learning}.
\newblock \bibinfo{journal}{\emph{arXiv preprint arXiv:1509.02971}}
  (\bibinfo{year}{2015}).
\newblock


\bibitem[Luo et~al\mbox{.}(2019)]%
        {luo2019real}
\bibfield{author}{\bibinfo{person}{Jiahuan Luo}, \bibinfo{person}{Xueyang Wu},
  \bibinfo{person}{Yun Luo}, \bibinfo{person}{Anbu Huang},
  \bibinfo{person}{Yunfeng Huang}, \bibinfo{person}{Yang Liu}, {and}
  \bibinfo{person}{Qiang Yang}.} \bibinfo{year}{2019}\natexlab{}.
\newblock \showarticletitle{Real-world image datasets for federated learning}.
\newblock \bibinfo{journal}{\emph{arXiv preprint arXiv:1910.11089}}
  (\bibinfo{year}{2019}).
\newblock


\bibitem[McMahan et~al\mbox{.}(2017)]%
        {fedavg_mcmahan2017communication}
\bibfield{author}{\bibinfo{person}{Brendan McMahan}, \bibinfo{person}{Eider
  Moore}, \bibinfo{person}{Daniel Ramage}, \bibinfo{person}{Seth Hampson},
  {and} \bibinfo{person}{Blaise~Aguera y Arcas}.}
  \bibinfo{year}{2017}\natexlab{}.
\newblock \showarticletitle{Communication-efficient learning of deep networks
  from decentralized data}. In \bibinfo{booktitle}{\emph{Artificial
  intelligence and statistics}}. PMLR, \bibinfo{pages}{1273--1282}.
\newblock


\bibitem[Sattler et~al\mbox{.}(2020)]%
        {2020_Sattler}
\bibfield{author}{\bibinfo{person}{Felix Sattler}, \bibinfo{person}{Simon
  Wiedemann}, \bibinfo{person}{Klaus-Robert Müller}, {and}
  \bibinfo{person}{Wojciech Samek}.} \bibinfo{year}{2020}\natexlab{}.
\newblock \showarticletitle{Robust and Communication-Efficient Federated
  Learning From Non-i.i.d. Data}.
\newblock \bibinfo{journal}{\emph{IEEE Transactions on Neural Networks and
  Learning Systems}} \bibinfo{volume}{31}, \bibinfo{number}{9}
  (\bibinfo{year}{2020}), \bibinfo{pages}{3400--3413}.
\newblock
\urldef\tempurl%
\url{https://doi.org/10.1109/TNNLS.2019.2944481}
\showDOI{\tempurl}


\bibitem[Simonyan et~al\mbox{.}(2015)]%
        {SimonyanZ14a}
\bibfield{author}{\bibinfo{person}{Karen Simonyan} {et~al\mbox{.}}}
  \bibinfo{year}{2015}\natexlab{}.
\newblock \showarticletitle{{Very Deep Convolutional Networks for Large-Scale
  Image Recognition}}. In \bibinfo{booktitle}{\emph{{ICLR} 2015}}.
\newblock


\bibitem[Sutton and Barto(2018)]%
        {sutton2018reinforcement}
\bibfield{author}{\bibinfo{person}{Richard~S Sutton} {and}
  \bibinfo{person}{Andrew~G Barto}.} \bibinfo{year}{2018}\natexlab{}.
\newblock \bibinfo{booktitle}{\emph{Reinforcement learning: An introduction}}.
\newblock \bibinfo{publisher}{MIT press}.
\newblock


\bibitem[Wang et~al\mbox{.}(2020)]%
        {favor_hwangInforcom2020}
\bibfield{author}{\bibinfo{person}{Hao Wang}, \bibinfo{person}{Zakhary Kaplan},
  \bibinfo{person}{Di Niu}, {and} \bibinfo{person}{Baochun Li}.}
  \bibinfo{year}{2020}\natexlab{}.
\newblock \showarticletitle{Optimizing Federated Learning on Non-IID Data with
  Reinforcement Learning}. In \bibinfo{booktitle}{\emph{IEEE INFOCOM 2020 -
  IEEE Conference on Computer Communications}}. \bibinfo{pages}{1698--1707}.
\newblock
\urldef\tempurl%
\url{https://doi.org/10.1109/INFOCOM41043.2020.9155494}
\showDOI{\tempurl}


\bibitem[Wang et~al\mbox{.}(2021b)]%
        {fednova}
\bibfield{author}{\bibinfo{person}{Jianyu Wang}, \bibinfo{person}{Qinghua Liu},
  \bibinfo{person}{Hao Liang}, \bibinfo{person}{Gauri Joshi}, {and}
  \bibinfo{person}{H.~Vincent Poor}.} \bibinfo{year}{2021}\natexlab{b}.
\newblock \showarticletitle{A Novel Framework for the Analysis and Design of
  Heterogeneous Federated Learning}.
\newblock \bibinfo{journal}{\emph{IEEE Transactions on Signal Processing}}
  \bibinfo{volume}{69} (\bibinfo{year}{2021}), \bibinfo{pages}{5234--5249}.
\newblock
\urldef\tempurl%
\url{https://doi.org/10.1109/TSP.2021.3106104}
\showDOI{\tempurl}


\bibitem[Wang et~al\mbox{.}(2021a)]%
        {ijcai2021_Fairness_fedfv}
\bibfield{author}{\bibinfo{person}{Zheng Wang}, \bibinfo{person}{Xiaoliang
  Fan}, \bibinfo{person}{Jianzhong Qi}, \bibinfo{person}{Chenglu Wen},
  \bibinfo{person}{Cheng Wang}, {and} \bibinfo{person}{Rongshan Yu}.}
  \bibinfo{year}{2021}\natexlab{a}.
\newblock \showarticletitle{Federated Learning with Fair Averaging}. In
  \bibinfo{booktitle}{\emph{Proceedings of the Thirtieth International Joint
  Conference on Artificial Intelligence, {IJCAI-21}}},
  \bibfield{editor}{\bibinfo{person}{Zhi-Hua Zhou}} (Ed.).
  \bibinfo{publisher}{International Joint Conferences on Artificial
  Intelligence Organization}, \bibinfo{pages}{1615--1623}.
\newblock
\newblock
\shownote{Main Track}.


\bibitem[Wang et~al\mbox{.}(2021c)]%
        {IWQOS2021_fedacs}
\bibfield{author}{\bibinfo{person}{Zibo Wang}, \bibinfo{person}{Yifei Zhu},
  \bibinfo{person}{Dan Wang}, {and} \bibinfo{person}{Zhu Han}.}
  \bibinfo{year}{2021}\natexlab{c}.
\newblock \showarticletitle{FedACS: Federated Skewness Analytics in
  Heterogeneous Decentralized Data Environments}. In
  \bibinfo{booktitle}{\emph{2021 IEEE/ACM 29th International Symposium on
  Quality of Service (IWQOS)}}. \bibinfo{pages}{1--10}.
\newblock


\bibitem[Wu and Wang(2021)]%
        {fedadp}
\bibfield{author}{\bibinfo{person}{Hongda Wu} {and} \bibinfo{person}{Ping
  Wang}.} \bibinfo{year}{2021}\natexlab{}.
\newblock \showarticletitle{Fast-Convergent Federated Learning With Adaptive
  Weighting}.
\newblock \bibinfo{journal}{\emph{IEEE Transactions on Cognitive Communications
  and Networking}} \bibinfo{volume}{7}, \bibinfo{number}{4}
  (\bibinfo{year}{2021}), \bibinfo{pages}{1078--1088}.
\newblock
\urldef\tempurl%
\url{https://doi.org/10.1109/TCCN.2021.3084406}
\showDOI{\tempurl}


\bibitem[Xiao et~al\mbox{.}(2017)]%
        {xiao2017fashion}
\bibfield{author}{\bibinfo{person}{Han Xiao}, \bibinfo{person}{Kashif Rasul},
  {and} \bibinfo{person}{Roland Vollgraf}.} \bibinfo{year}{2017}\natexlab{}.
\newblock \showarticletitle{Fashion-mnist: a novel image dataset for
  benchmarking machine learning algorithms}.
\newblock \bibinfo{journal}{\emph{arXiv preprint arXiv:1708.07747}}
  (\bibinfo{year}{2017}).
\newblock


\bibitem[Xiao et~al\mbox{.}(2020)]%
        {xiao2020averaging}
\bibfield{author}{\bibinfo{person}{Peng Xiao}, \bibinfo{person}{Samuel Cheng},
  \bibinfo{person}{Vladimir Stankovic}, {and} \bibinfo{person}{Dejan
  Vukobratovic}.} \bibinfo{year}{2020}\natexlab{}.
\newblock \showarticletitle{Averaging is probably not the optimum way of
  aggregating parameters in federated learning}.
\newblock \bibinfo{journal}{\emph{Entropy}} \bibinfo{volume}{22},
  \bibinfo{number}{3} (\bibinfo{year}{2020}), \bibinfo{pages}{314}.
\newblock


\bibitem[Yang(2021)]%
        {ijcai2021_HFL}
\bibfield{author}{\bibinfo{person}{He Yang}.} \bibinfo{year}{2021}\natexlab{}.
\newblock \showarticletitle{{ H-FL: A Hierarchical Communication-Efficient and
  Privacy-Protected Architecture for Federated Learning}}. In
  \bibinfo{booktitle}{\emph{Proceedings of the Thirtieth International Joint
  Conference on Artificial Intelligence, {IJCAI-21}}}.
  \bibinfo{pages}{479--485}.
\newblock
\newblock
\shownote{Main Track}.


\bibitem[Zeng et~al\mbox{.}(2021)]%
        {ICPP2021_fedCav}
\bibfield{author}{\bibinfo{person}{Hui Zeng}, \bibinfo{person}{Tongqing Zhou},
  \bibinfo{person}{Yeting Guo}, \bibinfo{person}{Zhiping Cai}, {and}
  \bibinfo{person}{Fang Liu}.} \bibinfo{year}{2021}\natexlab{}.
\newblock \bibinfo{booktitle}{\emph{FedCav: Contribution-Aware Model
  Aggregation on Distributed Heterogeneous Data in Federated Learning}}.
\newblock \bibinfo{publisher}{Association for Computing Machinery},
  \bibinfo{address}{New York, NY, USA}.
\newblock


\bibitem[Zhang et~al\mbox{.}(2021)]%
        {inforcom2021_fedsens}
\bibfield{author}{\bibinfo{person}{Daniel~Yue Zhang}, \bibinfo{person}{Ziyi
  Kou}, {and} \bibinfo{person}{Dong Wang}.} \bibinfo{year}{2021}\natexlab{}.
\newblock \showarticletitle{FedSens: A Federated Learning Approach for Smart
  Health Sensing with Class Imbalance in Resource Constrained Edge Computing}.
  In \bibinfo{booktitle}{\emph{IEEE INFOCOM 2021 - IEEE Conference on Computer
  Communications}}. \bibinfo{pages}{1--10}.
\newblock


\bibitem[Zhu et~al\mbox{.}(2021)]%
        {zhu2021federated}
\bibfield{author}{\bibinfo{person}{Hangyu Zhu}, \bibinfo{person}{Jinjin Xu},
  \bibinfo{person}{Shiqing Liu}, {and} \bibinfo{person}{Yaochu Jin}.}
  \bibinfo{year}{2021}\natexlab{}.
\newblock \showarticletitle{Federated Learning on Non-IID Data: A Survey}.
\newblock \bibinfo{journal}{\emph{arXiv preprint arXiv:2106.06843}}
  (\bibinfo{year}{2021}).
\newblock


\end{thebibliography}
\balance



\end{document}